\documentclass{article}

\PassOptionsToPackage{numbers, compress}{natbib}

\usepackage[final]{neurips_2023}

\usepackage[utf8]{inputenc} 
\usepackage[T1]{fontenc}    
\usepackage{hyperref}       
\usepackage{url}            
\usepackage{booktabs}       
\usepackage{amsfonts}       
\usepackage{nicefrac}       
\usepackage{microtype}      
\usepackage{xcolor}         
\usepackage[acronym]{glossaries}
\usepackage{enumitem}
\usepackage{graphicx}
\usepackage{caption}
\usepackage{subcaption}
\usepackage{amsthm}
\usepackage{wrapfig}

\title{Towards Fast Stochastic Sampling in Diffusion Generative Models}

\author{Kushagra Pandey \thanks{Work partially done during an internship at Bosch Center for Artificial Intelligence} \\
Department of Computer Science\\
University of California, Irvine\\
\texttt{pandeyk1@uci.edu} \\
\And
Maja Rudolph \\
Bosch Center for Artificial Intelligence \\
\texttt{Maja.Rudolph@us.bosch.com} \\
\AND
Stephan Mandt \\
Department of Computer Science \\
University of California, Irvine \\
\texttt{mandt@uci.edu}
}

\newcommand{\tr}{\textcolor{red}}

\newacronym{NFE}{NFE}{network function evaluations}
\newacronym{PSLD}{PSLD}{Phase Space Langevin Diffusion}
\newacronym{NSE}{NSE}{Naive Symplectic Euler}
\newacronym{RSE}{RSE}{Reduced Symplectic Euler}
\newacronym{CSE}{CSE}{Conjugate Symplectic Euler}
\newacronym{NVV}{NVV}{Naive Velocity Verlet}
\newacronym{RVV}{RVV}{Reduced Velocity Verlet}
\newacronym{CVV}{CVV}{Conjugate Velocity Verlet}
\newacronym{NOBA}{NOBA}{Naive OBA}
\newacronym{ROBA}{ROBA}{Reduced OBA}
\newacronym{COBA}{COBA}{Conjugate OBA}
\newacronym{SPS}{SPS}{Splitting-based PSLD Sampler}
\newacronym{CSPS}{CSPS}{Conjugate Splitting-based PSLD Sampler}

\usepackage{amsmath,amsfonts,bm}

\def\eqref#1{equation~\ref{#1}}

\def\1{\bm{1}}

\def\eps{{\epsilon}}

\def\rvm{{\mathbf{m}}}

\def\rvs{{\mathbf{s}}}

\def\rvw{{\mathbf{w}}}
\def\rvx{{\mathbf{x}}}

\def\rvz{{\mathbf{z}}}

\def\vtheta{{\bm{\theta}}}

\def\vf{{\bm{f}}}
\def\vg{{\bm{g}}}

\def\vs{{\bm{s}}}

\def\mF{{\bm{F}}}
\def\mG{{\bm{G}}}

\def\mI{{\bm{I}}}

\def\mSigma{{\bm{\Sigma}}}

\DeclareMathAlphabet{\mathsfit}{\encodingdefault}{\sfdefault}{m}{sl}
\SetMathAlphabet{\mathsfit}{bold}{\encodingdefault}{\sfdefault}{bx}{n}

\newcommand{\R}{\mathbb{R}}

\begin{document}

\maketitle

\begin{abstract}
  Diffusion models suffer from slow sample generation at inference time.
  Despite recent efforts, improving the sampling efficiency of stochastic samplers for diffusion models remains a promising direction.
  We propose \textit{Splitting Integrators} for fast stochastic sampling in pre-trained diffusion models in augmented spaces. Commonly used in molecular dynamics, splitting-based integrators attempt to improve sampling efficiency by cleverly alternating between numerical updates involving the data, auxiliary, or noise variables. However, we show that a \textit{naive} application of splitting integrators is sub-optimal for fast sampling. Consequently, we propose several principled \textit{modifications} to naive splitting samplers for improving sampling efficiency and denote the resulting samplers as \textit{Reduced Splitting Integrators}. In the context of \gls{PSLD} [Pandey \& Mandt, 2023] on CIFAR-10, our stochastic sampler achieves an FID score of 2.36 in only 100 \gls{NFE} as compared to 2.63 for the best baselines.
\end{abstract}

\section{Introduction}
Diffusion models \citep{sohl2015deep, song2019generative, ho2020denoising, songscore} have demonstrated impressive performance on various tasks, such as image and video synthesis \citep{dhariwal2021diffusion, ho2022cascaded, rombach2022high, https://doi.org/10.48550/arxiv.2204.06125, sahariaphotorealistic, https://doi.org/10.48550/arxiv.2203.09481, hovideo, harvey2022flexible}, image super-resolution \citep{saharia2022image}, and audio and speech synthesis \citep{chenwavegrad, lam2021bddm}. However, we identify two problems with the current landscape for fast sampling in diffusion models.

 Firstly, while fast sample generation in diffusion models is an active area of research, recent advances have mostly focused on accelerating deterministic samplers \citep{songdenoising, lu2022dpm, zhang2023fast}. However, the latter does not achieve optimal sample quality even with a large \gls{NFE} budget and may require specialized training improvements \citep{karraselucidating}. In contrast, stochastic samplers usually achieve better sample quality but require a larger sampling budget than their deterministic counterparts.
 
 Secondly, most efforts to improve diffusion model sampling have focused on a specific family of models that perform diffusion in the data space \citep{songscore, karraselucidating}. However, recent work \citep{dockhornscore, pandey2023generative, singhal2023where} indicates that performing diffusion in a joint space, where the data space is \textit{augmented} with auxiliary variables, can improve sample quality and likelihood over data-space-only diffusion models. However, improving the sampling efficiency for augmented diffusion models is still underexplored but a promising avenue for further improvements. Therefore, a generic framework is required for faster sampling from a broader class of diffusion models.

\textbf{Problem Statement: Efficient Stochastic Sampling during Inference.} Our goal is to develop efficient stochastic samplers that are applicable to sampling from a broader class of diffusion models (for instance, augmented diffusion models like PSLD \citep{pandey2023generative}) and achieve high-fidelity samples, even when the NFE budget is greatly reduced, e.g., from 1000 to 100 or even 50. We evaluate the effectiveness of the proposed samplers in the context of \gls{PSLD} \citep{pandey2023generative} due to its strong empirical performance. However, the presented techniques are also applicable to other diffusion models.

\textbf{Contributions.} Taking inspiration from molecular dynamics \citep{alma991006381329705251}, we present \textit{Splitting Integrators} for efficient sampling in diffusion models. However, we show that their naive application can be sub-optimal for sampling efficiency. Therefore, based on local error analysis for numerical solvers \citep{10.5555/153158}, we present several \textit{improvements} to our naive schemes to achieve improved sample efficiency. We denote the resulting samplers as \textit{Reduced Splitting Integrators}.

\section{Background}
We provide relevant background on diffusion models and their augmented versions. 
\label{sec:background}
\noindent
Specifically, diffusion models assume that a continuous-time \textit{forward process} (with affine drift),
\begin{equation}
    d\rvz_t = \mF_t\rvz_t \, dt + \mG_t \, d\rvw_t, \quad t \in [0, T],
    \label{eqn:fwd_process}
\end{equation}
with a standard Wiener process $\rvw_t$, time-dependent matrix $\mF \colon [0, T] \to \R^{d \times d}$, and diffusion coefficient $\mG \colon [0, T] \to \R^{d \times d}$, converts data $\rvz_0 \in \R^d$ into noise.
A \textit{reverse } SDE specifies how data is generated from noise \citep{songscore, ANDERSON1982313},
\begin{equation}
    d \rvz_t = \left[\mF_t\rvz_t - \mG_t \mG_t^\top \nabla_{\rvx_t} \log p_t(\rvz_t)\right] \, dt + \mG_t d\bar \rvw_t,
    \label{eq:reverse_time_diffusion}
\end{equation}
which involves the \textit{score} ($\nabla_{\rvz_t} \log p_t(\rvz_t)$) of the marginal distribution over $\rvz_t$ at time $t$.
 The score is intractable to compute and is approximated using a parametric estimator $\vs_{\theta}(\rvz_t, t)$, trained using denoising score matching \citep{song2019generative, songscore, 6795935}.
Once the score has been learned, generating new data samples involves sampling noise from the stationary distribution of Eqn.~\ref{eqn:fwd_process} (typically an isotropic Gaussian) and numerically integrating Eqn.~\ref{eq:reverse_time_diffusion}, resulting in a stochastic sampler. Next, we highlight two classes of diffusion models.

\textbf{Non-Augmented Diffusions.} Many existing diffusion models are formulated purely in data space, i.e., $\rvz_t = \rvx_t \in \mathbb{R}^d$. One popular example is the \textit{Variance Preserving} (VP)-SDE \citep{songscore} with $\mF_t = -\frac{1}{2}\beta_t \mI_d, \mG_t = \sqrt{\beta_t}\mI_d$.
Recently, \citet{karraselucidating} instead propose a re-scaled process, with $\mF_t = \bm{0}_d, \mG_t = \sqrt{2\dot{\sigma}_t\sigma_t}\mI_d$ ,which allows for faster sampling during generation.
Here $\beta_t, \sigma_t \in \mathbb{R}$ define the noise schedule in their respective diffusion processes.

\textbf{Augmented Diffusions.} For augmented diffusions, the data (or position) space, $\rvx_t$, is coupled with \textit{auxiliary} (a.k.a momentum) variables, $\rvm_t$, and diffusion is performed in the joint space. For instance, \citet{pandey2023generative} propose \gls{PSLD}, where $\rvz_t = [\rvx_t, \rvm_t]^T \in \mathbb{R}^{2d}$. Moreover,
\begin{equation}
\label{eqn:psld}
    \mF_t = \left(\frac{\beta}{2}\begin{pmatrix} -\Gamma & M^{-1}\\ -1 & - \nu \end{pmatrix} \otimes \mI_d\right), \quad\quad \mG_t = \left(\begin{pmatrix} \sqrt{\Gamma\beta} &0 \\ 0 &\sqrt{M\nu\beta} \end{pmatrix} \otimes \mI_d\right),
\end{equation}
 where $\{\beta, \Gamma, \nu, M^{-1}\} \in \mathbb{R}$ are the SDE hyperparameters.
 Augmented diffusions have been shown to exhibit better sample quality with a faster generation process \citep{dockhornscore, pandey2023generative}, and better likelihood estimation \citep{singhal2023where} over their non-augmented counterparts. In this work, we focus on sample quality and, therefore, study the efficient samplers we develop in the \gls{PSLD} setting.

\section{Splitting Integrators for Fast Stochastic Sampling}
\label{sec:splitting_main}
Splitting integrators are commonly used to design symplectic numerical solvers for molecular dynamics systems \citep{alma991006381329705251}. However, their application for fast diffusion sampling is still underexplored \citep{dockhornscore}. The main intuition behind splitting integrators is to \textit{split} an ODE/SDE into subcomponents, which are then \textit{independently solved} numerically (or analytically). The resulting updates are then \textit{composed} in a specific order to obtain the final solution. We briefly introduce splitting integrators in Appendix \ref{app:split_intro} and refer interested readers to \citet{alma991006381329705251} for a detailed discussion. Splitting integrators are particularly suited for augmented diffusion models since they can leverage the split into position and momentum variables for faster sampling. However, their application for fast sampling in position-space-only diffusion models \citep{songscore, karraselucidating} remains an interesting direction for future work.

\begin{figure}
     \begin{subfigure}[b]{0.25\textwidth}
         \centering
         \includegraphics[width=\textwidth]{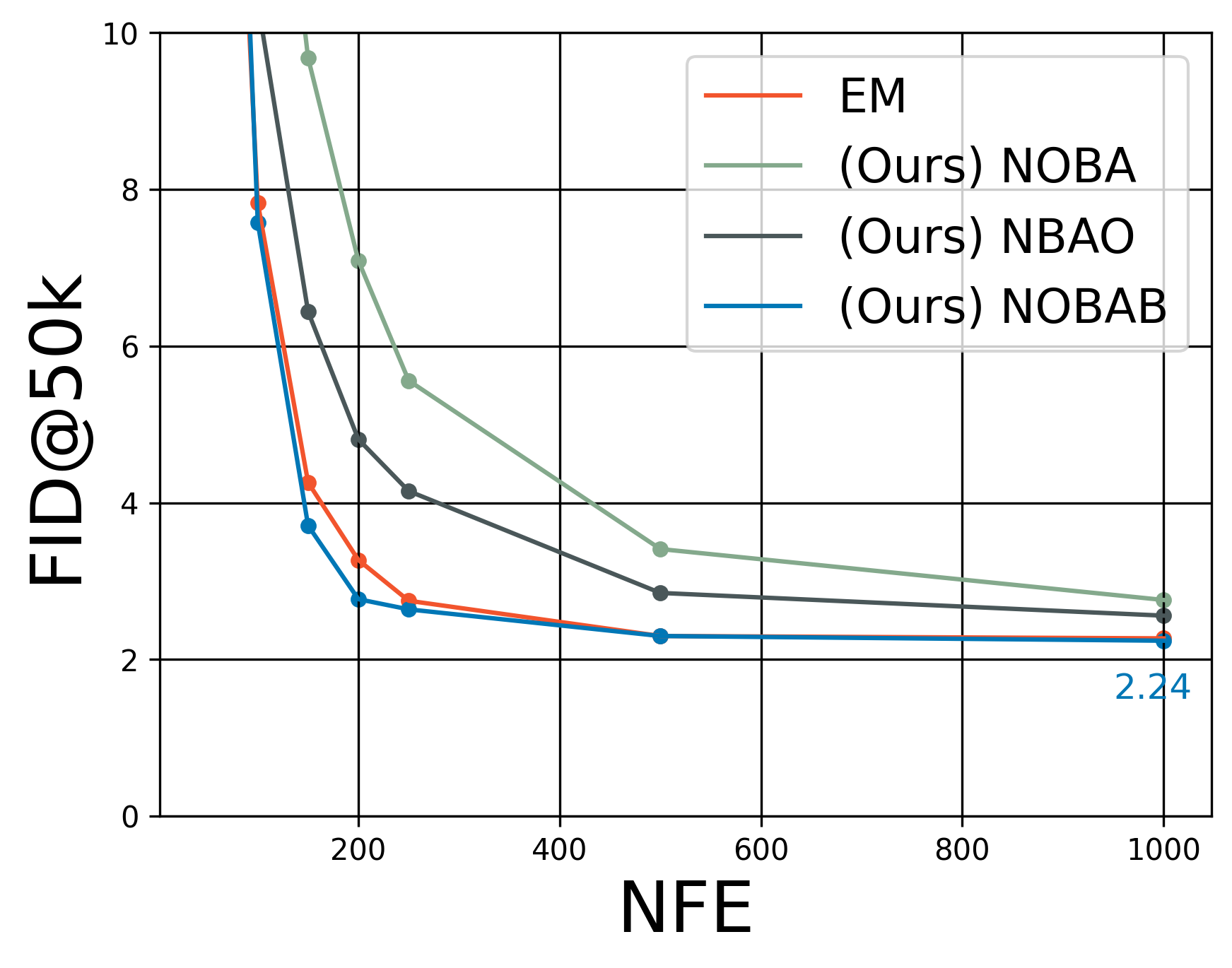}
         \caption{}
         \label{fig:w1a}
     \end{subfigure}
     \hfill
     \begin{subfigure}[b]{0.24\textwidth}
         \centering
         \includegraphics[width=\textwidth]{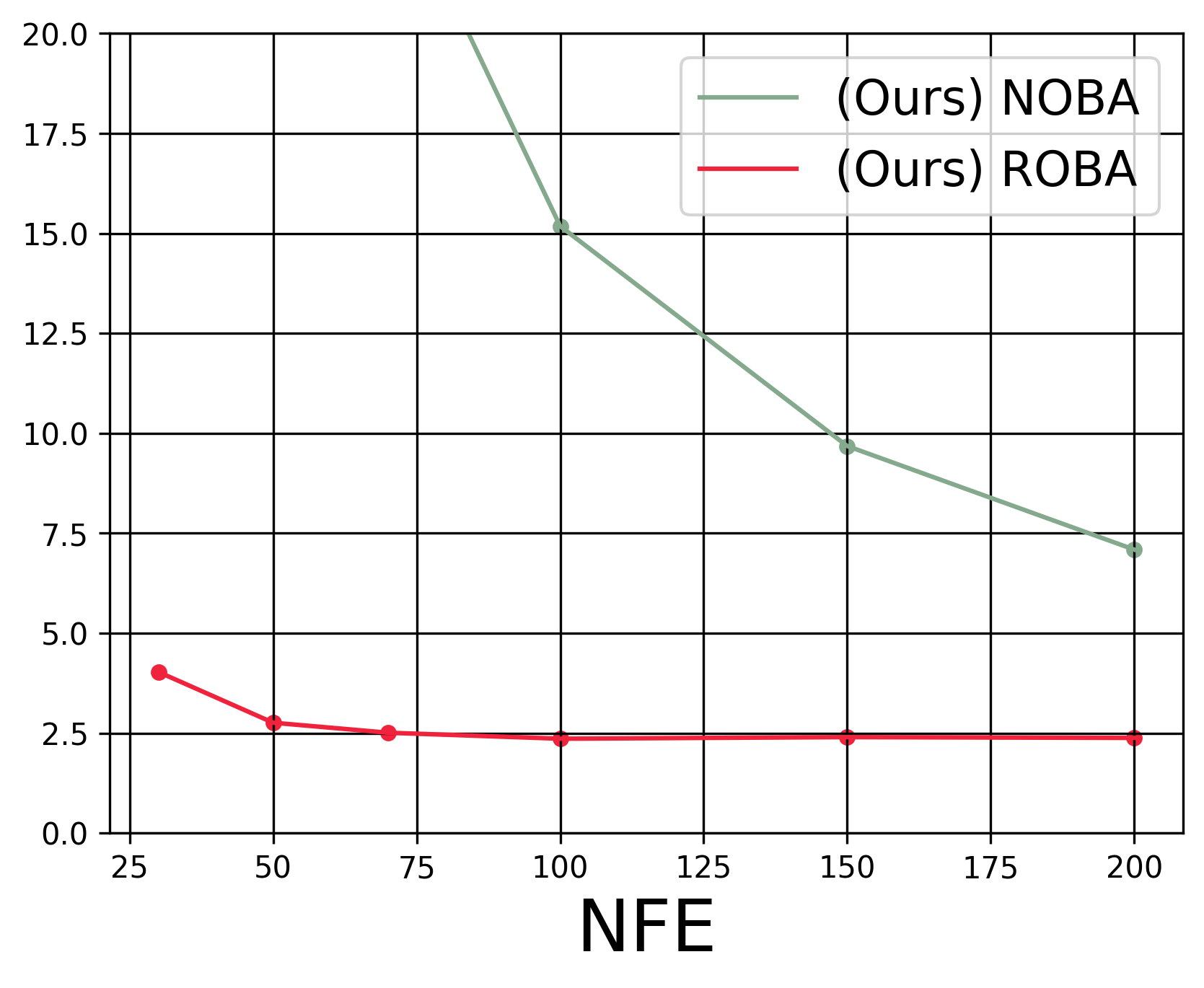}
         \caption{}
         \label{fig:w1b}
     \end{subfigure}
     \hfill
     \begin{subfigure}[b]{0.24\textwidth}
         \centering
         \includegraphics[width=\textwidth]{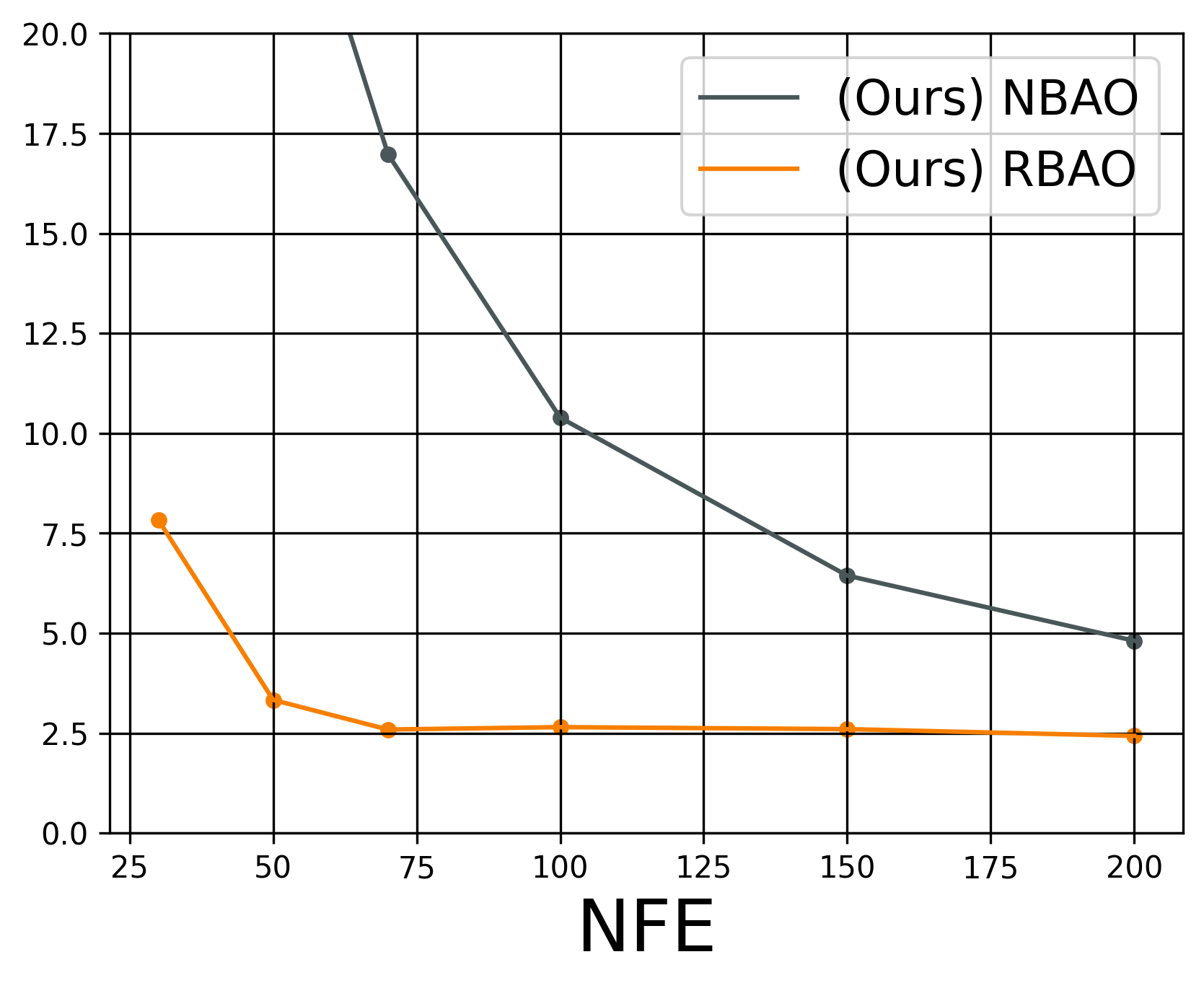}
         \caption{}
         \label{fig:w1c}
     \end{subfigure}
     \begin{subfigure}[b]{0.24\textwidth}
         \centering
         \includegraphics[width=\textwidth]{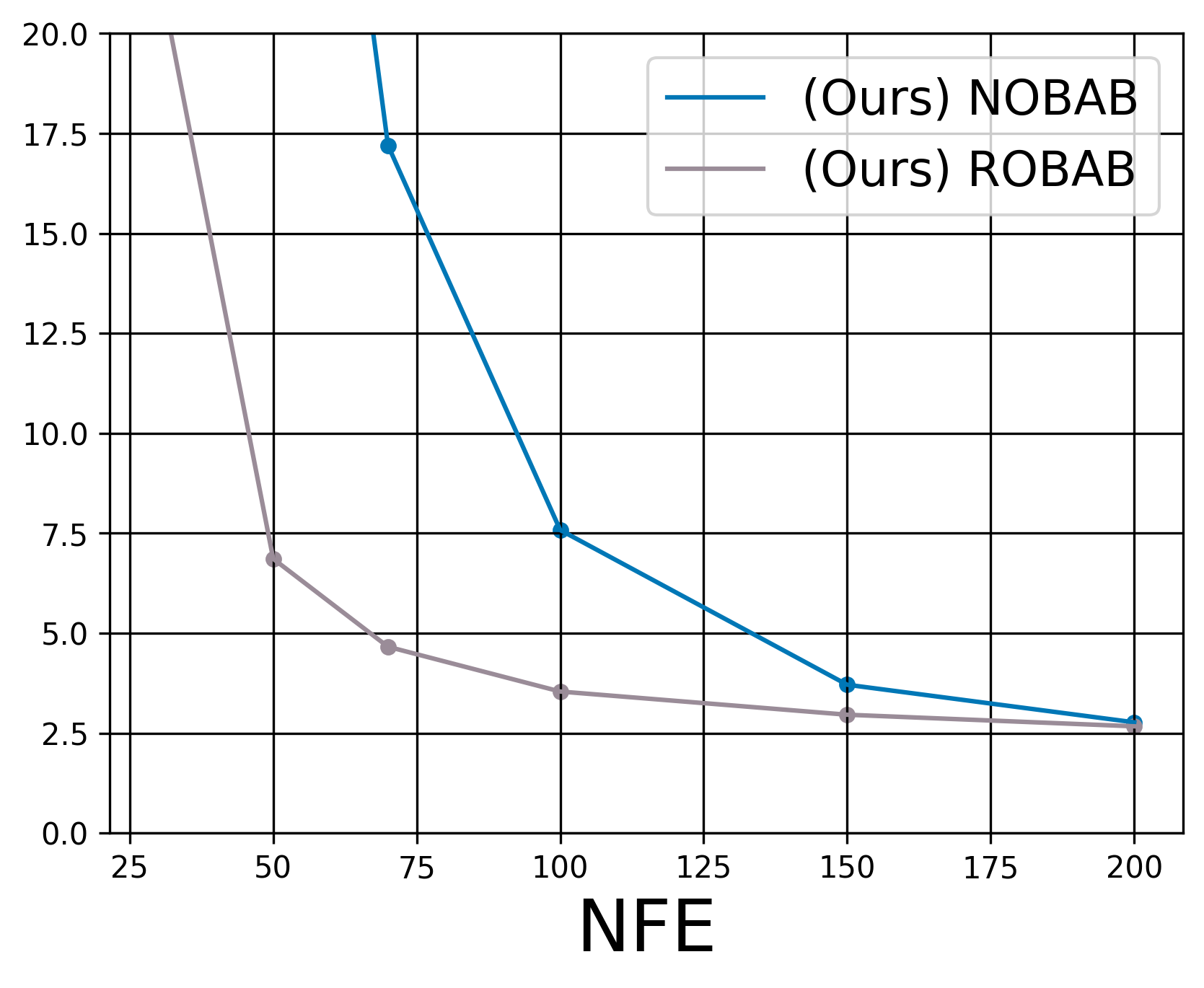}
         \caption{}
         \label{fig:w1d}
     \end{subfigure}
        \caption{Splitting integrators for fast stochastic sampling in diffusion models. a) Sample quality comparisons between naive splitting schemes and the EM Sampler. (b-d) Reduced Splitting schemes outperform their naive counterparts by a large margin. (Left to Right) Comparisons between the naive splitting schemes and their reduced counterparts for the OBA, BAO, and OBAB samplers.}
        \label{fig:fig_splitting}
\end{figure}

\textbf{Naive Stochastic Splitting Integrators.}
We apply splitting integrators to the \gls{PSLD} Reverse SDE and use the following splitting scheme.
\begin{equation*}
    \begin{pmatrix}d\bar{\rvx}_t \\ d\bar{\rvm}_t\end{pmatrix} = \underbrace{\frac{\beta}{2}\begin{pmatrix} 2\Gamma \bar{\rvx}_t - M^{-1} \bar{\rvm}_t + 2\Gamma \vs_{\theta}^x(\bar{\rvz}_t, t)\\ 0 \end{pmatrix} dt}_{A} + O + \underbrace{\frac{\beta}{2}\begin{pmatrix} 0\\ \bar{\rvx}_t +2\nu\bar{\rvm}_t + 2M\nu \vs_{\theta}^m(\bar{\rvz}_t, t) \end{pmatrix} dt}_{B}.
\end{equation*}
where $\bar{\rvx}_{t} = \rvx_{T-t}$, $\bar{\rvm}_{t} = \rvm_{T-t}$, $\vs_{\vtheta}^x$ and $\vs_{\vtheta}^m$ denote the score components in the data and momentum space, respectively, and $O = \begin{pmatrix} -\frac{\beta\Gamma}{2} \bar{\rvx}_t dt + \sqrt{\beta\Gamma} d\bar{\rvw}_t \\ -\frac{\beta\nu}{2} \bar{\rvm}_t dt + \sqrt{M\nu\beta} d\bar{\rvw}_t \end{pmatrix}$ represents the Ornstein-Uhlenbeck process in the joint space. Given step size $h$, we further denote the Euler updates for the components $A$, $B$ and $O$ as $\mathcal{L}_h^A$, $\mathcal{L}_h^B$ and $\mathcal{L}_h^O$, respectively. These updates can be composed differently to yield a valid stochastic sampler. Among several possible composition schemes, we find OBA, BAO, and OBAB to work particularly well (See Appendix \ref{sec:split_stochastic} for complete numerical updates). We refer to these schemes as \textit{Naive Splitting Integrators}. Consequently, we denote these samplers as NOBA, NBAO, and NOBAB, respectively.
For an empirical evaluation, we utilize a \gls{PSLD} model \citep{pandey2023generative} ($\Gamma=0.01, \nu=4.01$) pre-trained on CIFAR-10. We measure sampling efficiency via network function evaluations (\gls{NFE}) and measure sample quality using FID \citep{heusel2017gans}.
Our proposed naive samplers outperform the baseline Euler-Maruyama (EM) sampler (Fig. \ref{fig:w1a}). 
This is intuitive since, unlike EM, the proposed naive samplers alternate between updates in the momentum and the position space, thus exploiting the coupling between the data and the momentum variables. Next, we propose several modifications to the naive schemes to further improve sampling efficiency.

\textbf{Reduced Stochastic Splitting Integrators.}
While the proposed \textit{naive} schemes perform well, Fig. \ref{fig:w1a} also suggests scope for improvement, especially at low \gls{NFE} budgets. This suggests the need for a deeper insight into the error analysis for the naive schemes. Therefore, based on local error analysis for SDEs, we propose the following \textit{modifications} to our naive samplers.
\begin{itemize}[leftmargin=*]
    \item Firstly, we reuse the score function evaluation between the first consecutive position and the momentum updates. 
    \item Secondly, for sampling schemes involving half-steps (i.e. with step-size $h/2$ like NOBAB), we evaluate the score in the last step with a timestep embedding of $T - (t + h)$ instead of $T-t$.
    \item Lastly, similar to \citet{karraselucidating}, we introduce a parameter $\lambda_s$ in the position space update for $\mathcal{L}_O$ to control the amount of noise injected in the position space. However, adding a similar parameter in the momentum space led to unstable behavior. Therefore, we restrict this adjustment to the position space. For a given sampling budget, we explicitly adjust $\lambda_s$ for optimal sample quality via a simple grid search during inference. We include more details in Appendix \ref{sec:effects_sto}
\end{itemize}
Applying these modifications to naive splitting integrators yields \textit{Reduced Splitting Integrators}. Consequently, we denote these samplers as ROBA, RBAO, and ROBAB, respectively (See Appendix \ref{sec:sto_adj_updates} for complete numerical updates). Figures \ref{fig:w1b}-\ref{fig:w1d} compare the performance of our reduced samplers with their naive counterparts. Empirically, our modifications significantly improve sample quality for all three samplers (Fig. \ref{fig:w1b}-\ref{fig:w1d}).
This is because our proposed modifications serve the following benefits.

\begin{itemize}[leftmargin=*]
    \item First, re-using the score function evaluation
    between the first consecutive updates in the data and the momentum space 
    reduces the number of \gls{NFE}s per update step by one for all reduced samplers, which enables smaller step sizes for the same compute budget during sampling and hence largely reduces numerical discretization errors.
    \item Secondly, re-using the score function evaluation leads to canceling certain error terms arising from numerical discretization, which is especially helpful for low \gls{NFE} budgets.
    \item Lastly, as also noted in recent works \citep{karraselucidating, bao2022analyticdpm}, the choice of the variance of the noise injected during stochastic sampling can largely affect sample quality. 
\end{itemize}
\section{Additional Experimental Results}
\label{sec:sota}

\begin{figure}
     \begin{subfigure}[b]{0.3\textwidth}
         \centering
         \includegraphics[width=\textwidth]{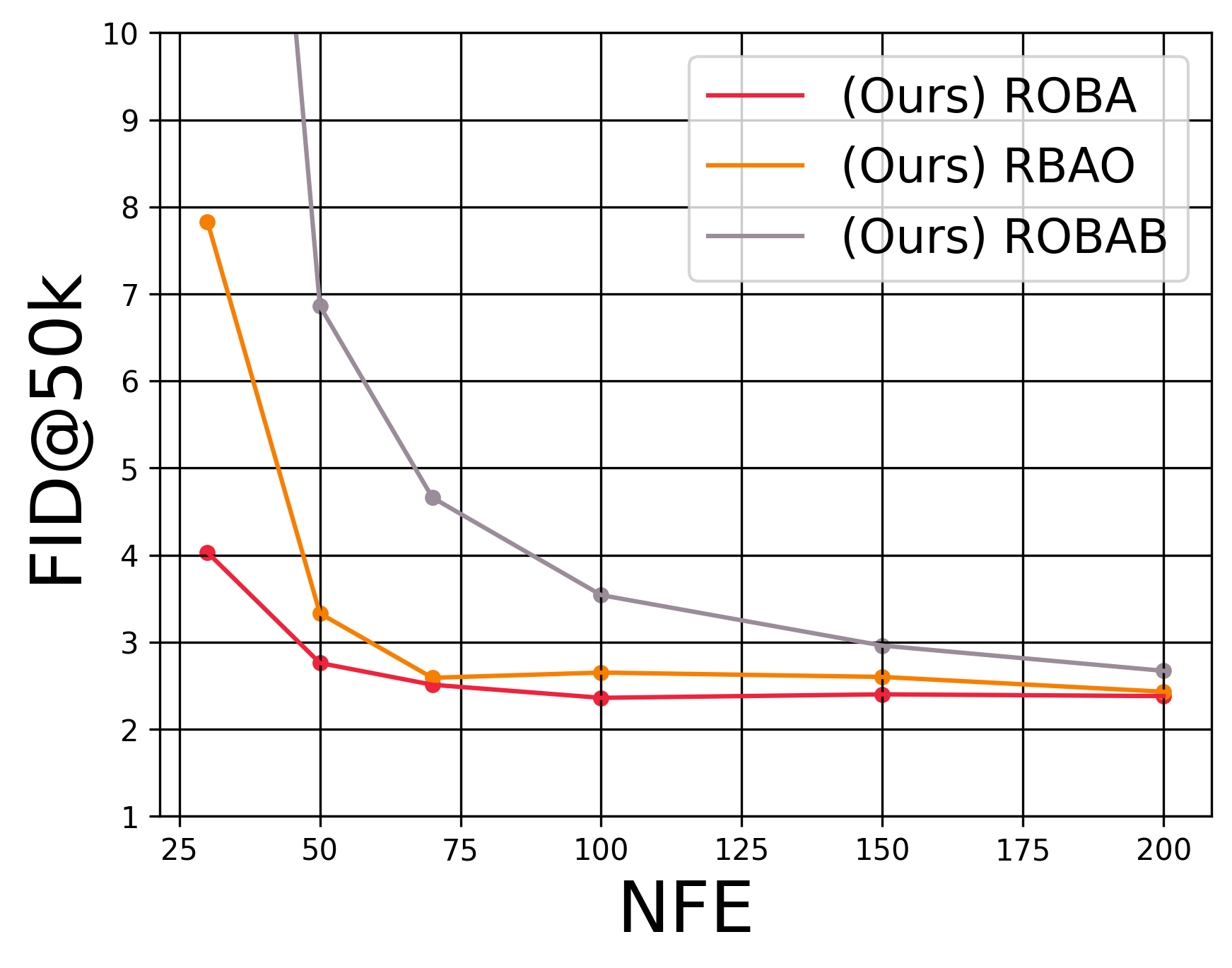}
         \caption{}
         \label{fig:w2a}
     \end{subfigure}
     \hfill
     \begin{subfigure}[b]{0.29\textwidth}
         \centering
         \includegraphics[width=\textwidth]{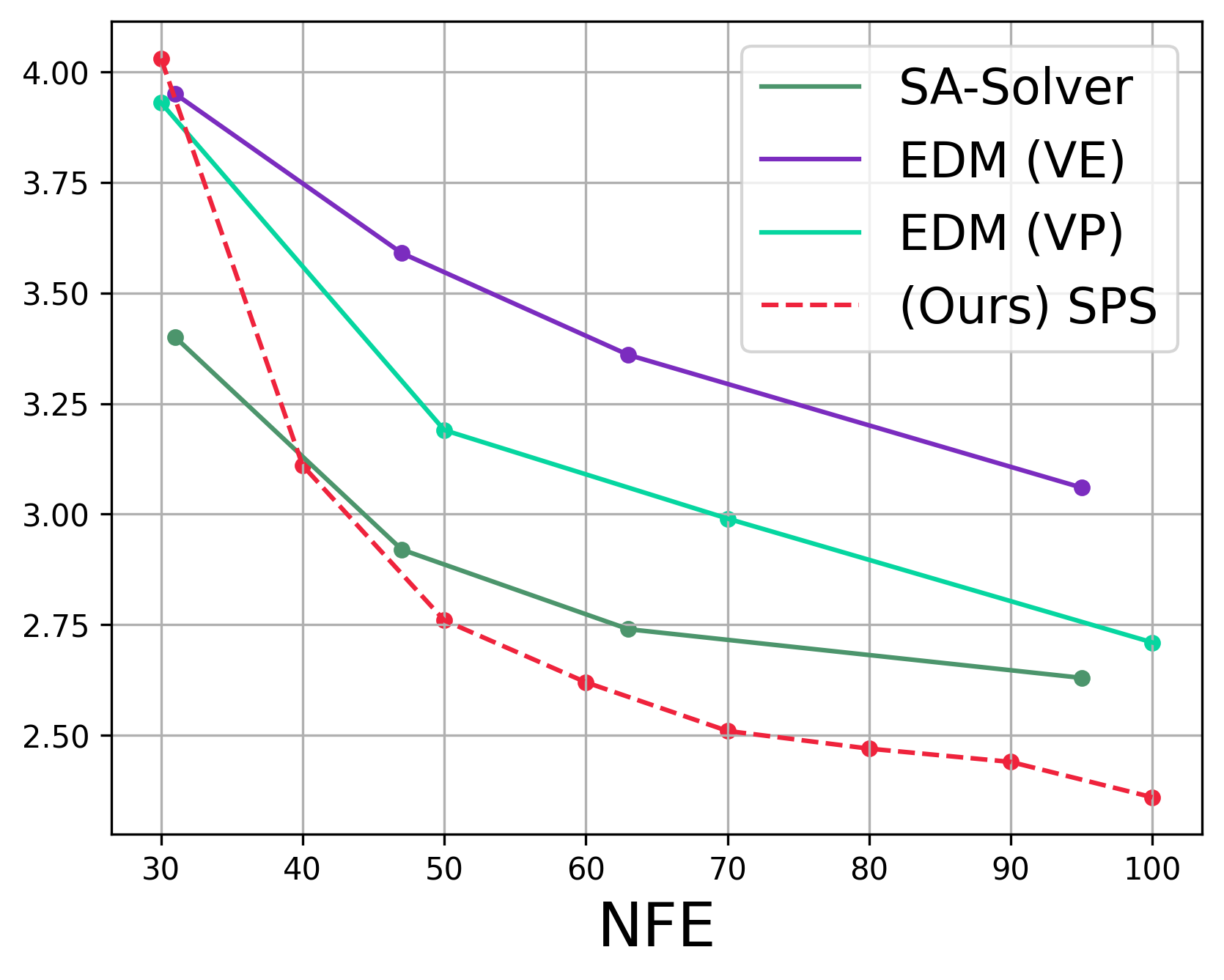}
         \caption{}
         \label{fig:sota_cifar10}
     \end{subfigure}
     \hfill
     \begin{subfigure}[b]{0.285\textwidth}
         \centering
         \includegraphics[width=\textwidth]{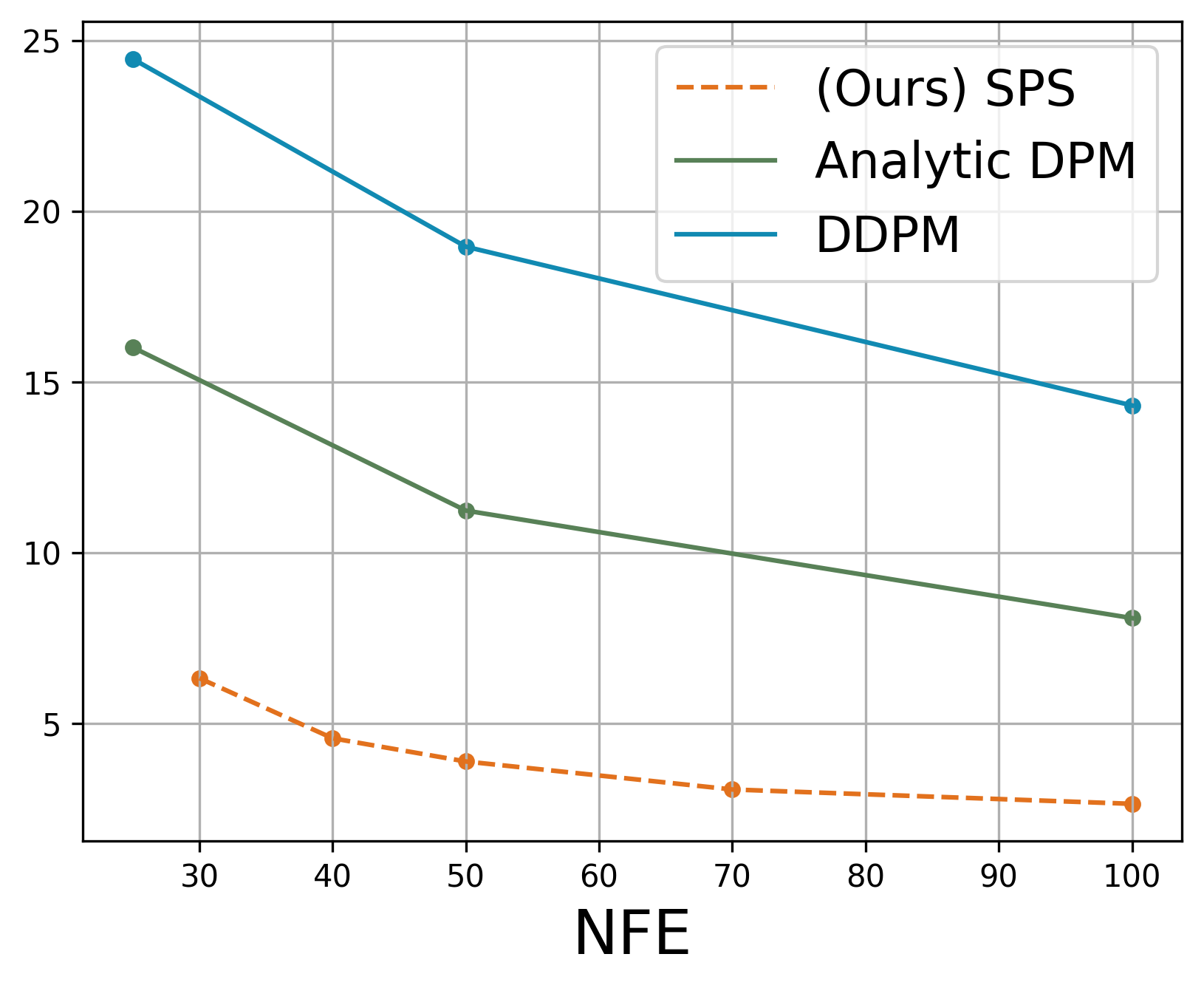}
         \caption{}
         \label{fig:sota_celeba64}
     \end{subfigure}
    \caption{Comparison with state-of-the-art methods. a) Reduced OBA outperforms other reduced splitting schemes. (b, c) SPS outperforms competing baseline methods for stochastic sampling for the CIFAR-10 and CelebA-64 datasets, respectively.}
    \label{fig:fig_sota}
\end{figure}

\textbf{Datasets and Evaluation Metrics.} We use the CIFAR-10 and the CelebA-64 \citep{liu2015faceattributes} datasets for comparisons. Unless specified otherwise, we report FID for 50k generated samples for all datasets and quantify sampling efficiency using \gls{NFE}. We use the pre-trained models from PSLD \citep{pandey2023generative} and use a last-step denoising step \citep{jolicoeur-martineau2021adversarial, songscore} when sampling from the proposed samplers.

\textbf{Baselines and setup.} While the techniques presented in this work are generally applicable to other types of diffusion models, we compare our best-performing reduced splitting integrator, ROBA (see Fig. \ref{fig:w2a}), with recent work on fast stochastic sampling in diffusion models including SA-Solver \citep{xue2023sasolver}, SEEDS \citep{gonzalez2023seeds}, EDM \citep{karraselucidating} and Analytic DPM \citep{bao2022analyticdpm}. We also compare with another splitting-based sampler, SSCS \citep{dockhornscore}, applied to the PSLD reverse SDE. We denote our best-performing sampler as \emph{\gls{SPS}} for notational convenience.

\textbf{Empirical Results.} For CIFAR-10, our SDE sampler outperforms all other baselines for different compute budgets (Fig. \ref{fig:sota_cifar10}). We make similar observations for the CelebA-64 dataset, where our proposed samplers can obtain significant gains over prior methods (see Fig. \ref{fig:sota_celeba64}). More specifically, our SDE sampler achieves an FID score of \textbf{2.36} and \textbf{2.64} for the CIFAR-10 and CelebA-64 datasets, respectively, in just 100 \acrlong{NFE}. We include additional results in Appendix \ref{app:add_exps}.
\section{Conclusion}
We present a splitting integrator for fast stochastic sampling from a broader class of diffusion models. We show that a naive application of splitting integrators can be sub-optimal for sample quality. To this end, we propose a few modifications to our naive samplers and denote the resulting sampler as Reduced Splitting Integrators, which outperform existing baselines for fast stochastic sampling. Empirically, we find that controlling the amount of stochasticity injected in the data space during sampling can largely affect sample quality for the proposed splitting integrators. Therefore, further investigation into the theoretical aspects of optimal noise injection in stochastic sampling can be an interesting direction for future work. We also believe that exploring similar splitting samplers for position-space-only diffusion models like Stable-Diffusion \citep{rombach2022high} could also be an interesting direction for further work.

\bibliographystyle{unsrtnat}
\bibliography{neurips_2023}

\begin{thebibliography}{35}
\providecommand{\natexlab}[1]{#1}
\providecommand{\url}[1]{\texttt{#1}}
\expandafter\ifx\csname urlstyle\endcsname\relax
  \providecommand{\doi}[1]{doi: #1}\else
  \providecommand{\doi}{doi: \begingroup \urlstyle{rm}\Url}\fi

\bibitem[Sohl-Dickstein et~al.(2015)Sohl-Dickstein, Weiss, Maheswaranathan, and Ganguli]{sohl2015deep}
Jascha Sohl-Dickstein, Eric Weiss, Niru Maheswaranathan, and Surya Ganguli.
\newblock Deep unsupervised learning using nonequilibrium thermodynamics.
\newblock In \emph{International Conference on Machine Learning}, pages 2256--2265. PMLR, 2015.

\bibitem[Song and Ermon(2019)]{song2019generative}
Yang Song and Stefano Ermon.
\newblock Generative modeling by estimating gradients of the data distribution.
\newblock \emph{Advances in neural information processing systems}, 32, 2019.

\bibitem[Ho et~al.(2020)Ho, Jain, and Abbeel]{ho2020denoising}
Jonathan Ho, Ajay Jain, and Pieter Abbeel.
\newblock Denoising diffusion probabilistic models.
\newblock \emph{Advances in Neural Information Processing Systems}, 33:\penalty0 6840--6851, 2020.

\bibitem[Song et~al.(2020)Song, Sohl-Dickstein, Kingma, Kumar, Ermon, and Poole]{songscore}
Yang Song, Jascha Sohl-Dickstein, Diederik~P Kingma, Abhishek Kumar, Stefano Ermon, and Ben Poole.
\newblock Score-based generative modeling through stochastic differential equations.
\newblock In \emph{International Conference on Learning Representations}, 2020.

\bibitem[Dhariwal and Nichol(2021)]{dhariwal2021diffusion}
Prafulla Dhariwal and Alexander Nichol.
\newblock Diffusion models beat gans on image synthesis.
\newblock \emph{Advances in Neural Information Processing Systems}, 34:\penalty0 8780--8794, 2021.

\bibitem[Ho et~al.(2022{\natexlab{a}})Ho, Saharia, Chan, Fleet, Norouzi, and Salimans]{ho2022cascaded}
Jonathan Ho, Chitwan Saharia, William Chan, David~J Fleet, Mohammad Norouzi, and Tim Salimans.
\newblock Cascaded diffusion models for high fidelity image generation.
\newblock \emph{J. Mach. Learn. Res.}, 23\penalty0 (47):\penalty0 1--33, 2022{\natexlab{a}}.

\bibitem[Rombach et~al.(2022)Rombach, Blattmann, Lorenz, Esser, and Ommer]{rombach2022high}
Robin Rombach, Andreas Blattmann, Dominik Lorenz, Patrick Esser, and Bj{\"o}rn Ommer.
\newblock High-resolution image synthesis with latent diffusion models.
\newblock In \emph{Proceedings of the IEEE/CVF Conference on Computer Vision and Pattern Recognition}, pages 10684--10695, 2022.

\bibitem[Ramesh et~al.(2022)Ramesh, Dhariwal, Nichol, Chu, and Chen]{https://doi.org/10.48550/arxiv.2204.06125}
Aditya Ramesh, Prafulla Dhariwal, Alex Nichol, Casey Chu, and Mark Chen.
\newblock Hierarchical text-conditional image generation with clip latents, 2022.
\newblock URL \url{https://arxiv.org/abs/2204.06125}.

\bibitem[Saharia et~al.(2022{\natexlab{a}})Saharia, Chan, Saxena, Li, Whang, Denton, Ghasemipour, Gontijo~Lopes, Karagol~Ayan, Salimans, et~al.]{sahariaphotorealistic}
Chitwan Saharia, William Chan, Saurabh Saxena, Lala Li, Jay Whang, Emily~L Denton, Kamyar Ghasemipour, Raphael Gontijo~Lopes, Burcu Karagol~Ayan, Tim Salimans, et~al.
\newblock Photorealistic text-to-image diffusion models with deep language understanding.
\newblock volume~35, pages 36479--36494, 2022{\natexlab{a}}.

\bibitem[Yang et~al.(2022)Yang, Srivastava, and Mandt]{https://doi.org/10.48550/arxiv.2203.09481}
Ruihan Yang, Prakhar Srivastava, and Stephan Mandt.
\newblock Diffusion probabilistic modeling for video generation, 2022.
\newblock URL \url{https://arxiv.org/abs/2203.09481}.

\bibitem[Ho et~al.(2022{\natexlab{b}})Ho, Salimans, Gritsenko, Chan, Norouzi, and Fleet]{hovideo}
Jonathan Ho, Tim Salimans, Alexey~A. Gritsenko, William Chan, Mohammad Norouzi, and David~J. Fleet.
\newblock Video diffusion models.
\newblock In \emph{ICLR Workshop on Deep Generative Models for Highly Structured Data}, 2022{\natexlab{b}}.
\newblock URL \url{https://openreview.net/forum?id=BBelR2NdDZ5}.

\bibitem[Harvey et~al.(2022)Harvey, Naderiparizi, Masrani, Weilbach, and Wood]{harvey2022flexible}
William Harvey, Saeid Naderiparizi, Vaden Masrani, Christian Weilbach, and Frank Wood.
\newblock Flexible diffusion modeling of long videos.
\newblock \emph{Advances in Neural Information Processing Systems}, 35:\penalty0 27953--27965, 2022.

\bibitem[Saharia et~al.(2022{\natexlab{b}})Saharia, Ho, Chan, Salimans, Fleet, and Norouzi]{saharia2022image}
Chitwan Saharia, Jonathan Ho, William Chan, Tim Salimans, David~J Fleet, and Mohammad Norouzi.
\newblock Image super-resolution via iterative refinement.
\newblock \emph{IEEE Transactions on Pattern Analysis and Machine Intelligence}, 2022{\natexlab{b}}.

\bibitem[Chen et~al.(2021)Chen, Zhang, Zen, Weiss, Norouzi, and Chan]{chenwavegrad}
Nanxin Chen, Yu~Zhang, Heiga Zen, Ron~J Weiss, Mohammad Norouzi, and William Chan.
\newblock Wavegrad: Estimating gradients for waveform generation.
\newblock In \emph{International Conference on Learning Representations}, 2021.
\newblock URL \url{https://openreview.net/forum?id=NsMLjcFaO8O}.

\bibitem[Lam et~al.(2021)Lam, Wang, Su, and Yu]{lam2021bddm}
Max~WY Lam, Jun Wang, Dan Su, and Dong Yu.
\newblock Bddm: Bilateral denoising diffusion models for fast and high-quality speech synthesis.
\newblock In \emph{International Conference on Learning Representations}, 2021.

\bibitem[Song et~al.(2021)Song, Meng, and Ermon]{songdenoising}
Jiaming Song, Chenlin Meng, and Stefano Ermon.
\newblock Denoising diffusion implicit models.
\newblock In \emph{International Conference on Learning Representations}, 2021.
\newblock URL \url{https://openreview.net/forum?id=St1giarCHLP}.

\bibitem[Lu et~al.(2022)Lu, Zhou, Bao, Chen, Li, and Zhu]{lu2022dpm}
Cheng Lu, Yuhao Zhou, Fan Bao, Jianfei Chen, Chongxuan Li, and Jun Zhu.
\newblock Dpm-solver: A fast ode solver for diffusion probabilistic model sampling in around 10 steps.
\newblock \emph{Advances in Neural Information Processing Systems}, 35:\penalty0 5775--5787, 2022.

\bibitem[Zhang and Chen(2023)]{zhang2023fast}
Qinsheng Zhang and Yongxin Chen.
\newblock Fast sampling of diffusion models with exponential integrator.
\newblock In \emph{The Eleventh International Conference on Learning Representations}, 2023.
\newblock URL \url{https://openreview.net/forum?id=Loek7hfb46P}.

\bibitem[Karras et~al.(2022)Karras, Aittala, Aila, and Laine]{karraselucidating}
Tero Karras, Miika Aittala, Timo Aila, and Samuli Laine.
\newblock Elucidating the design space of diffusion-based generative models.
\newblock \emph{Advances in Neural Information Processing Systems}, 35:\penalty0 26565--26577, 2022.

\bibitem[Dockhorn et~al.(2022)Dockhorn, Vahdat, and Kreis]{dockhornscore}
Tim Dockhorn, Arash Vahdat, and Karsten Kreis.
\newblock Score-based generative modeling with critically-damped langevin diffusion.
\newblock In \emph{International Conference on Learning Representations}, 2022.
\newblock URL \url{https://openreview.net/forum?id=CzceR82CYc}.

\bibitem[Pandey and Mandt(2023)]{pandey2023generative}
Kushagra Pandey and Stephan Mandt.
\newblock A complete recipe for diffusion generative models.
\newblock In \emph{Proceedings of the IEEE/CVF International Conference on Computer Vision (ICCV)}, pages 4261--4272, October 2023.

\bibitem[Singhal et~al.(2023)Singhal, Goldstein, and Ranganath]{singhal2023where}
Raghav Singhal, Mark Goldstein, and Rajesh Ranganath.
\newblock Where to diffuse, how to diffuse, and how to get back: Automated learning for multivariate diffusions.
\newblock In \emph{The Eleventh International Conference on Learning Representations}, 2023.
\newblock URL \url{https://openreview.net/forum?id=osei3IzUia}.

\bibitem[Leimkuhler(2015)]{alma991006381329705251}
B.~Leimkuhler.
\newblock \emph{Molecular dynamics : with deterministic and stochastic numerical methods / Ben Leimkuhler, Charles Matthews.}
\newblock Interdisciplinary applied mathematics, 39. Springer, Cham, 2015.
\newblock ISBN 3319163744.

\bibitem[Hairer et~al.(1993)Hairer, N\o{}rsett, and Wanner]{10.5555/153158}
E.~Hairer, S.~P. N\o{}rsett, and G.~Wanner.
\newblock \emph{Solving Ordinary Differential Equations I (2nd Revised. Ed.): Nonstiff Problems}.
\newblock Springer-Verlag, Berlin, Heidelberg, 1993.
\newblock ISBN 0387566708.

\bibitem[Anderson(1982)]{ANDERSON1982313}
Brian~D.O. Anderson.
\newblock Reverse-time diffusion equation models.
\newblock \emph{Stochastic Processes and their Applications}, 12\penalty0 (3):\penalty0 313--326, 1982.
\newblock ISSN 0304-4149.
\newblock \doi{https://doi.org/10.1016/0304-4149(82)90051-5}.
\newblock URL \url{https://www.sciencedirect.com/science/article/pii/0304414982900515}.

\bibitem[Vincent(2011)]{6795935}
Pascal Vincent.
\newblock A connection between score matching and denoising autoencoders.
\newblock \emph{Neural Computation}, 23\penalty0 (7):\penalty0 1661--1674, 2011.
\newblock \doi{10.1162/NECO_a_00142}.

\bibitem[Heusel et~al.(2017)Heusel, Ramsauer, Unterthiner, Nessler, and Hochreiter]{heusel2017gans}
Martin Heusel, Hubert Ramsauer, Thomas Unterthiner, Bernhard Nessler, and Sepp Hochreiter.
\newblock Gans trained by a two time-scale update rule converge to a local nash equilibrium.
\newblock \emph{Advances in neural information processing systems}, 30, 2017.

\bibitem[Bao et~al.(2022)Bao, Li, Zhu, and Zhang]{bao2022analyticdpm}
Fan Bao, Chongxuan Li, Jun Zhu, and Bo~Zhang.
\newblock Analytic-{DPM}: an analytic estimate of the optimal reverse variance in diffusion probabilistic models.
\newblock In \emph{International Conference on Learning Representations}, 2022.
\newblock URL \url{https://openreview.net/forum?id=0xiJLKH-ufZ}.

\bibitem[Liu et~al.(2015)Liu, Luo, Wang, and Tang]{liu2015faceattributes}
Ziwei Liu, Ping Luo, Xiaogang Wang, and Xiaoou Tang.
\newblock Deep learning face attributes in the wild.
\newblock In \emph{Proceedings of International Conference on Computer Vision (ICCV)}, December 2015.

\bibitem[Jolicoeur-Martineau et~al.(2021)Jolicoeur-Martineau, Pich{\'e}-Taillefer, Mitliagkas, and des Combes]{jolicoeur-martineau2021adversarial}
Alexia Jolicoeur-Martineau, R{\'e}mi Pich{\'e}-Taillefer, Ioannis Mitliagkas, and Remi~Tachet des Combes.
\newblock Adversarial score matching and improved sampling for image generation.
\newblock In \emph{International Conference on Learning Representations}, 2021.
\newblock URL \url{https://openreview.net/forum?id=eLfqMl3z3lq}.

\bibitem[Xue et~al.(2023)Xue, Yi, Luo, Zhang, Sun, Li, and Ma]{xue2023sasolver}
Shuchen Xue, Mingyang Yi, Weijian Luo, Shifeng Zhang, Jiacheng Sun, Zhenguo Li, and Zhi-Ming Ma.
\newblock Sa-solver: Stochastic adams solver for fast sampling of diffusion models, 2023.

\bibitem[Gonzalez et~al.(2023)Gonzalez, Fernandez, Tran, Gherbi, Hajri, and Masmoudi]{gonzalez2023seeds}
Martin Gonzalez, Nelson Fernandez, Thuy Tran, Elies Gherbi, Hatem Hajri, and Nader Masmoudi.
\newblock Seeds: Exponential sde solvers for fast high-quality sampling from diffusion models, 2023.

\bibitem[Kloeden and Platen(1992)]{Kloeden1992}
Peter~E. Kloeden and Eckhard Platen.
\newblock \emph{Numerical Solution of Stochastic Differential Equations}.
\newblock Springer Berlin Heidelberg, 1992.
\newblock \doi{10.1007/978-3-662-12616-5}.
\newblock URL \url{https://doi.org/10.1007/978-3-662-12616-5}.

\bibitem[Krizhevsky(2009)]{krizhevsky2009learning}
Alex Krizhevsky.
\newblock Learning multiple layers of features from tiny images.
\newblock pages 32--33, 2009.
\newblock URL \url{https://www.cs.toronto.edu/~kriz/learning-features-2009-TR.pdf}.

\bibitem[Obukhov et~al.(2020)Obukhov, Seitzer, Wu, Zhydenko, Kyl, and Lin]{obukhov2020torchfidelity}
Anton Obukhov, Maximilian Seitzer, Po-Wei Wu, Semen Zhydenko, Jonathan Kyl, and Elvis Yu-Jing Lin.
\newblock High-fidelity performance metrics for generative models in pytorch, 2020.
\newblock URL \url{https://github.com/toshas/torch-fidelity}.
\newblock Version: 0.3.0, DOI: 10.5281/zenodo.4957738.

\end{thebibliography}

\appendix
\section{Splitting Integrators}
\subsection{Introduction to Splitting Integrators}
\label{app:split_intro}
Here, we provide a brief introduction to splitting integrators. For a detailed account of splitting integrators for designing symplectic numerical methods, we refer interested readers to \citet{alma991006381329705251}. As discussed in the main text, the main idea behind splitting integrators is to split the vector field of an ODE or the drift and the diffusion components of an SDE into independent subcomponents, which are then solved independently using a numerical scheme (or analytically). The solutions to independent sub-components are then composed in a specific order to obtain the final solution. Thus, three key steps in designing a splitting integrator are \textbf{split}, \textbf{solve}, and \textbf{compose}. We illustrate these steps with an example of a deterministic dynamical system. However, the concept is generic and can be applied to systems with stochastic dynamics as well. 

Consider a dynamical system specified by the following ODE:
\begin{equation}
    \begin{pmatrix}d\rvx_t \\ d\rvm_t\end{pmatrix} = \begin{pmatrix}
        \vf(\rvx_t, \rvm_t) \\ \vg(\rvx_t, \rvm_t)
    \end{pmatrix}dt
    \label{eqn:app_split_1}
\end{equation}

We start by choosing a scheme to split the vector field for the ODE in Eqn. \ref{eqn:app_split_1}. While different types of splitting schemes can be possible, we choose the following scheme for this example,
\begin{equation}
    \begin{pmatrix}d\rvx_t \\ d\rvm_t\end{pmatrix} = \underbrace{\begin{pmatrix}
        \vf(\rvx_t, \rvm_t) \\ 0
    \end{pmatrix}dt}_{A} + \underbrace{\begin{pmatrix}
        0 \\ \vg(\rvx_t, \rvm_t)
    \end{pmatrix}dt}_{B}
\end{equation}
where we denote the individual components by $A$ and $B$. Next, we solve each of these components independently, i.e., we compute solutions for the following ODEs independently.
\begin{equation}
    \begin{pmatrix}d\rvx_t \\ d\rvm_t\end{pmatrix} = \begin{pmatrix}
        \vf(\rvx_t, \rvm_t) \\ 0
    \end{pmatrix}dt, \quad\quad \begin{pmatrix}d\rvx_t \\ d\rvm_t\end{pmatrix} = \begin{pmatrix}
        0 \\ \vg(\rvx_t, \rvm_t)
    \end{pmatrix}dt
\end{equation}
While any numerical scheme can be used to approximate the solution for the splitting components, we use Euler throughout this work. Therefore, applying an Euler approximation, with a step size $h$, to each of these splitting components yields the solutions $\mathcal{L}^A_h$ and $\mathcal{L}^B_h$, as follows,
\begin{equation}
    \mathcal{L}^A_h = \begin{cases}
        \rvx_{t+h} = \rvx_t + h \vf(\rvx_t, \rvm_t) \\
        \rvm_{t+h} = \rvm_t
    \end{cases}, \quad \mathcal{L}^B_h = \begin{cases}
        \rvx_{t+h} = \rvx_t \\
        \rvm_{t+h} = \rvm_t + h \vg(\rvx_t, \rvm_t)
    \end{cases}
\end{equation}
In the final step, we compose the solutions to the independent components in a specific order. For instance, for the composition scheme AB, the final solution $\mathcal{L}^{[AB]}_h = \mathcal{L}^B_h \circ \mathcal{L}^A_h$. Therefore,
\begin{equation}
    \mathcal{L}^{[AB]}_h = \begin{cases}
        \rvx_{t+h} = \rvx_t + h \vf(\rvx_t, \rvm_t) \\
        \rvm_{t+h} = \rvm_t + h \vg(\rvx_{t+h}, \rvm_t)
    \end{cases}
\end{equation}
is the required solution. It is worth noting that the final solution depends on the chosen composition scheme, and often it is not clear beforehand which composition scheme might work best.

\subsection{Stochastic Splitting Integrators}
\label{sec:split_stochastic}
We split the Reverse Diffusion SDE for PSLD using the following splitting scheme.
\begin{equation}
    \begin{pmatrix}d\bar{\rvx}_t \\ d\bar{\rvm}_t\end{pmatrix} = \underbrace{\frac{\beta}{2}\begin{pmatrix} 2\Gamma \bar{\rvx}_t - M^{-1} \bar{\rvm}_t + 2\Gamma \vs_{\theta}^x(\bar{\rvz}_t, t)\\ 0 \end{pmatrix} dt}_{A} + O + \underbrace{\frac{\beta}{2}\begin{pmatrix} 0\\ \bar{\rvx}_t +2\nu\bar{\rvm}_t + 2M\nu \vs_{\theta}^m(\bar{\rvz}_t, t) \end{pmatrix} dt}_{B}
\end{equation}
where $O = \begin{pmatrix} -\frac{\beta\Gamma}{2} \bar{\rvx}_t dt + \sqrt{\beta\Gamma} d\bar{\rvw}_t \\ -\frac{\beta\nu}{2} \bar{\rvm}_t dt + \sqrt{M\nu\beta} d\bar{\rvw}_t \end{pmatrix}$ is the Ornstein-Uhlenbeck component which injects stochasticity during sampling. Similar to the deterministic case, $\bar{\rvx}_{t} = \rvx_{T-t}$, $\bar{\rvm}_{t} = \rvm_{T-t}$, $\vs_{\vtheta}^x$ and $\vs_{\vtheta}^m$ denote the score components in the data and momentum space, respectively. We approximate the solution for splits $A$ and $B$ using a simple Euler-based numerical approximation. Formally, we denote the Euler approximation for the splits $A$ and $B$ by $\mathcal{L}_A$ and $\mathcal{L}_B$, respectively, with their corresponding numerical updates specified as:
\begin{align}
    \mathcal{L}_A &: \begin{cases}
        \bar{\rvx}_{t+h} &= \bar{\rvx}_{t} + \frac{h\beta}{2}\Big[2\Gamma \bar{\rvx}_t - M^{-1} \bar{\rvm}_t + 2\Gamma \vs_{\theta}^x(\bar{\rvx}_t, \bar{\rvm}_t, T-t)\Big] \\
        \bar{\rvm}_{t+h} &= \bar{\rvm}_{t}
    \end{cases} \\
    \mathcal{L}_B &: \begin{cases}
        \bar{\rvx}_{t+h} &= \bar{\rvx}_{t} \\
        \bar{\rvm}_{t+h} &= \bar{\rvm}_{t} + \frac{h\beta}{2}\Big[\bar{\rvx}_t +2\nu\bar{\rvm}_t + 2M\nu \vs_{\theta}^m(\bar{\rvx}_t, \bar{\rvm}_t, T-t)\Big]
    \end{cases}
\end{align}
It is worth noting that the solution to the OU component can be computed analytically:
\begin{align}
\mathcal{L}_O: \begin{cases}
    \bar{\rvx}_{t+h} &= \exp{\left(\frac{-h\beta\Gamma}{2}\right)}\bar{\rvx}_t + \sqrt{1 - \exp{\left(-h\beta\Gamma\right)}}\bm{\epsilon}_x, \;\;\;\; \bm{\epsilon}_x \sim \mathcal{N}(\bm{0}_d, \mI_d) \\
\bar{\rvm}_{t+h} &= \exp{\left(\frac{-h\beta\nu}{2}\right)}\bar{\rvm}_t + \sqrt{M}\sqrt{1 - \exp{\left(-h\beta\nu\right)}}\bm{\epsilon}_m, \;\;\;\; \bm{\epsilon}_m \sim \mathcal{N}(\bm{0}_d, \mI_d)
\end{cases}
\end{align}

Next, we highlight the numerical update equations for the Naive OBA, BAO, and OBAB samplers and their corresponding reduced analogues.

\subsubsection{Naive Splitting Samplers}
\textbf{Naive OBA}: In this scheme, for a given step size h, the solutions to the splitting pieces $\mathcal{L}_h^A$, $\mathcal{L}_h^B$ and $\mathcal{L}_h^O$ are composed as $\mathcal{L}^{[OBA]}_h = \mathcal{L}^A_h \circ \mathcal{L}^B_h \circ\mathcal{L}^O_h$.
Consequently, one numerical update step for this integrator can be defined as,
\begin{align}
    \bar{\rvx}_{t+h} &= \exp{\left(\frac{-h\beta\Gamma}{2}\right)}\bar{\rvx}_t + \sqrt{1 - \exp{\left(-h\beta\Gamma\right)}}\bm{\epsilon}_x \\
    \bar{\rvm}_{t+h} &= \exp{\left(\frac{-h\beta\nu}{2}\right)}\bar{\rvm}_t + \sqrt{M}\sqrt{1 - \exp{\left(-h\beta\nu\right)}}\bm{\epsilon}_m \\
    \hat{\rvm}_{t+h} &= \bar{\rvm}_{t+h} + \frac{h\beta}{2}\left[\bar{\rvx}_{t+h} +2\nu\bar{\rvm}_{t+h} + 2M\nu \vs_{\theta}^m(\bar{\rvx}_{t+h}, \bar{\rvm}_{t+h}, T - t)\right] \\
    \hat{\rvx}_{t+h} &= \bar{\rvx}_{t+h} + \frac{h\beta}{2}\left[2\Gamma \bar{\rvx}_{t+h} - M^{-1} \hat{\rvm}_{t+h} + 2\Gamma \vs_{\theta}^x(\bar{\rvx}_{t+h}, \hat{\rvm}_{t+h}, T - t)\right]
\end{align}
where $\bm{\epsilon}_x, \bm{\epsilon}_m \sim \mathcal{N}(\bm{0}_d, \mI_d)$. Therefore, one update step for Naive OBA requires \textbf{two NFEs}.

\textbf{Naive BAO}: Given a step size h, the solutions to the splitting pieces $\mathcal{L}_h^A$, $\mathcal{L}_h^B$ and $\mathcal{L}_h^O$ are composed as $\mathcal{L}^{[BAO]}_h = \mathcal{L}^O_h \circ \mathcal{L}^A_h \circ\mathcal{L}^B_h$.
Consequently, one numerical update step for this integrator can be defined as,
\begin{align}
    \bar{\rvm}_{t+h} &= \bar{\rvm}_t + \frac{h\beta}{2}\left[\bar{\rvx}_{t} +2\nu\bar{\rvm}_{t} + 2M\nu \vs_{\theta}^m(\bar{\rvx}_{t}, \bar{\rvm}_{t}, T - t)\right] \\
    \bar{\rvx}_{t+h} &= \bar{\rvx}_t + \frac{h\beta}{2}\left[2\Gamma \bar{\rvx}_t - M^{-1} \bar{\rvm}_{t+h} + 2\Gamma \vs_{\theta}^x(\bar{\rvx}_t, \bar{\rvm}_{t+h}, T - t)\right] \\
    \hat{\rvx}_{t+h} &= \exp{\left(\frac{-h\beta\Gamma}{2}\right)}\bar{\rvx}_{t+h} + \sqrt{1 - \exp{\left(-h\beta\Gamma\right)}}\bm{\epsilon}_x \\
    \hat{\rvm}_{t+h} &= \exp{\left(\frac{-h\beta\nu}{2}\right)}\bar{\rvm}_{t+h} + \sqrt{M}\sqrt{1 - \exp{\left(-h\beta\nu\right)}}\bm{\epsilon}_m
\end{align}
where $\bm{\epsilon}_x, \bm{\epsilon}_m \sim \mathcal{N}(\bm{0}_d, \mI_d)$. Therefore, one update step for Naive BAO requires \textbf{two NFEs}.

\textbf{Naive OBAB}: Given a step size h, the solutions to the splitting pieces $\mathcal{L}_h^A$, $\mathcal{L}_h^B$ and $\mathcal{L}_h^O$ are composed as $\mathcal{L}^{[OBAB]}_h = \mathcal{L}^B_{h/2} \circ \mathcal{L}^A_h \circ \mathcal{L}^B_{h/2} \circ\mathcal{L}^O_h$.
Consequently, one numerical update step for this integrator can be defined as,
\begin{align}
    \bar{\rvx}_{t+h} &= \exp{\left(\frac{-h\beta\Gamma}{2}\right)}\bar{\rvx}_t + \sqrt{1 - \exp{\left(-h\beta\Gamma\right)}}\bm{\epsilon}_x \\
    \bar{\rvm}_{t+h} &= \exp{\left(\frac{-h\beta\nu}{2}\right)}\bar{\rvm}_t + \sqrt{M}\sqrt{1 - \exp{\left(-h\beta\nu\right)}}\bm{\epsilon}_m \\
    \hat{\rvm}_{t+h/2} &= \bar{\rvm}_{t+h} + \frac{h\beta}{4}\left[\bar{\rvx}_{t+h} +2\nu\bar{\rvm}_{t+h} + 2M\nu \vs_{\theta}^m(\bar{\rvx}_{t+h}, \bar{\rvm}_{t+h}, T - t)\right]
\end{align}
\begin{align}
    \hat{\rvx}_{t+h} &= \bar{\rvx}_{t+h} + \frac{h\beta}{2}\left[2\Gamma \bar{\rvx}_{t+h} - M^{-1} \hat{\rvm}_{t+h/2} + 2\Gamma \vs_{\theta}^x(\bar{\rvx}_{t+h}, \hat{\rvm}_{t+h/2}, T - t)\right] \\
    \hat{\rvm}_{t+h} &= \hat{\rvm}_{t+h/2} + \frac{h\beta}{4}\left[\hat{\rvx}_{t+h} +2\nu\hat{\rvm}_{t+h/2} + 2M\nu \vs_{\theta}^m(\hat{\rvx}_{t+h}, \hat{\rvm}_{t+h/2}, T - t)\right]
\end{align}
where $\bm{\epsilon}_x, \bm{\epsilon}_m \sim \mathcal{N}(\bm{0}_d, \mI_d)$. Therefore, one update step for Naive OBAB requires \textbf{three NFEs}.

\subsubsection{Effects of controlling stochasticity}
\label{sec:effects_sto}
Similar to \citet{karraselucidating}, we introduce a parameter $\lambda_s$ in the position space update for $\mathcal{L}_O$ to control the amount of noise injected in the position space. More specifically, we modify the numerical update equations for the Ornstein-Uhlenbeck process in the position space as follows:
\begin{equation}
    \bar{\rvx}_{t+h} = \exp{\left(\frac{-h\beta\Gamma}{2}\right)}\bar{\rvx}_t + \sqrt{1 - \exp{\left(-\bar{t}\lambda_s\beta\Gamma\right)}}\bm{\epsilon}_x, \;\;\;\; \bm{\epsilon}_x \sim \mathcal{N}(\bm{0}_d, \mI_d)
\end{equation}
where $\bar{t} = \frac{(T - t) + (T - t - h)}{2}$, i.e., the mid-point for two consecutive time steps during sampling. Our choice of $\bar{t}$ is primarily based on empirical results. Moreover, we also choose $\lambda_s$ empirically and tune it for a given step size h during inference. 
We also found that adding a similar noise scaling parameter in the momentum space led to unstable sampling. Therefore, we restrict this adjustment to only the position space.

\subsubsection{Reduced Splitting Schemes}
\label{sec:sto_adj_updates}
We obtain the Reduced Splitting schemes by sharing the score function evaluation between the first consecutive position and momentum updates for all samplers. Additionally, for half-step updates (as in the OBAB scheme), we condition the score function with the timestep embedding of $T - (t+h)$ instead of $T - t$. Moreover, we make the adjustments as described in Appendix \ref{sec:effects_sto}. The changes in \tr{red} indicate the differences between the naive and the reduced samplers.

\textbf{Reduced OBA}: The numerical updates for this scheme are as follows (the terms in \tr{red} denote the changes from the Naive OBA scheme),
\begin{align}
    \bar{\rvx}_{t+h} &= \exp{\left(\frac{-h\beta\Gamma}{2}\right)}\bar{\rvx}_t + \tr{\sqrt{1 - \exp{\left(-\bar{t}\lambda_s\beta\Gamma\right)}}\bm{\epsilon}_x} \\
    \bar{\rvm}_{t+h} &= \exp{\left(\frac{-h\beta\nu}{2}\right)}\bar{\rvm}_t + \sqrt{M}\sqrt{1 - \exp{\left(-h\beta\nu\right)}}\bm{\epsilon}_m \\
    \hat{\rvm}_{t+h} &= \bar{\rvm}_{t+h} + \frac{h\beta}{2}\left[\bar{\rvx}_{t+h} +2\nu\bar{\rvm}_{t+h} + 2M\nu \vs_{\theta}^m(\bar{\rvx}_{t+h}, \bar{\rvm}_{t+h}, T - t)\right] \\
    \hat{\rvx}_{t+h} &= \bar{\rvx}_{t+h} + \frac{h\beta}{2}\left[2\Gamma \bar{\rvx}_{t+h} - M^{-1} \hat{\rvm}_{t+h} + 2\Gamma \tr{\vs_{\theta}^x(\bar{\rvx}_{t+h}, \bar{\rvm}_{t+h}, T - t)}\right]
\end{align}
where $\bm{\epsilon}_x, \bm{\epsilon}_m \sim \mathcal{N}(\bm{0}_d, \mI_d)$. It is worth noting that Reduced OBA requires only \textbf{one NFE} per update step since a single score evaluation is re-used in both the momentum and the position updates.

\textbf{Reduced BAO}: The numerical updates for this scheme are as follows,
\begin{align}
    \bar{\rvm}_{t+h} &= \bar{\rvm}_t + \frac{h\beta}{2}\left[\bar{\rvx}_t +2\nu\bar{\rvm}_t + 2M\nu \vs_{\theta}^m(\bar{\rvx}_t, \bar{\rvm}_t, T - t)\right] \\
    \bar{\rvx}_{t+h} &= \bar{\rvx}_t + \frac{h\beta}{2}\left[2\Gamma \bar{\rvx}_t - M^{-1} \bar{\rvm}_{t+h} + 2\Gamma \tr{\vs_{\theta}^x(\bar{\rvx}_t, \bar{\rvm}_t, T - t)}\right]\\
    \hat{\rvx}_{t+h} &= \exp{\left(\frac{-h\beta\Gamma}{2}\right)}\bar{\rvx}_{t+h} + \tr{\sqrt{1 - \exp{\left(-\bar{t}\lambda_s\beta\Gamma\right)}}\bm{\epsilon}_x} \\
    \hat{\rvm}_{t+h} &= \exp{\left(\frac{-h\beta\nu}{2}\right)}\bar{\rvm}_{t+h} + \sqrt{M}\sqrt{1 - \exp{\left(-h\beta\nu\right)}}\bm{\epsilon}_m 
\end{align}
where $\bm{\epsilon}_x, \bm{\epsilon}_m \sim \mathcal{N}(\bm{0}_d, \mI_d)$. Similar to the Reduced OBA scheme, Reduced BAO also requires only \textbf{one NFE} per update step since a single score evaluation is re-used in both the momentum and the position updates.

\textbf{Reduced OBAB}: The numerical updates for this scheme are as follows,
\begin{align}
    \bar{\rvx}_{t+h} &= \exp{\left(\frac{-h\beta\Gamma}{2}\right)}\bar{\rvx}_t + \tr{\sqrt{1 - \exp{\left(-\bar{t}\lambda_s\beta\Gamma\right)}}\bm{\epsilon}_x} \\
    \bar{\rvm}_{t+h} &= \exp{\left(\frac{-h\beta\nu}{2}\right)}\bar{\rvm}_t + \sqrt{M}\sqrt{1 - \exp{\left(-h\beta\nu\right)}}\bm{\epsilon}_m \\
    \hat{\rvm}_{t+h/2} &= \bar{\rvm}_{t+h} + \frac{h\beta}{4}\left[\bar{\rvx}_{t+h} +2\nu\bar{\rvm}_{t+h} + 2M\nu \vs_{\theta}^m(\bar{\rvx}_{t+h}, \bar{\rvm}_{t+h}, T - t)\right] \\
    \hat{\rvx}_{t+h} &= \bar{\rvx}_{t+h} + \frac{h\beta}{2}\left[2\Gamma \bar{\rvx}_{t+h} - M^{-1} \hat{\rvm}_{t+h/2} + 2\Gamma \tr{\vs_{\theta}^x(\bar{\rvx}_{t+h}, \bar{\rvm}_{t+h}, T - t)}\right]\\
    \hat{\rvm}_{t+h} &= \hat{\rvm}_{t+h/2} + \frac{h\beta}{4}\left[\hat{\rvx}_{t+h} +2\nu\hat{\rvm}_{t+h/2} + 2M\nu \tr{\vs_{\theta}^m(\hat{\rvx}_{t+h}, \hat{\rvm}_{t+h/2}, T - (t + h))}\right]
\end{align}
where $\bm{\epsilon}_x, \bm{\epsilon}_m \sim \mathcal{N}(\bm{0}_d, \mI_d)$. It is worth noting that, in contrast to the Reduced OBA and BAO schemes, Reduced OBAB requires \textbf{two NFE} per update step.

\section{Local Error Analysis and Justification for design choices}

We now analyze the naive and reduced splitting samplers proposed in this work from the lens of local error analysis for SDE solvers. The probability flow SDE for PSLD is defined as,
\begin{equation}
    \begin{pmatrix}d\bar{\rvx}_t \\ d\bar{\rvm}_t\end{pmatrix} = \frac{\beta}{2}\begin{pmatrix} \Gamma \bar{\rvx}_t - M^{-1} \bar{\rvm}_t + 2\Gamma \vs_{\theta}^x(\bar{\rvz}_t, T-t)\\ \bar{\rvx}_t +\nu\bar{\rvm}_t + 2M\nu \vs_{\theta}^m(\bar{\rvz}_t, T-t) \end{pmatrix}dt, \qquad t \in [0, T]
\end{equation}
We denote the proposed numerical discretization schemes by $\mathcal{G}_h$ and the underlying ground-truth trajectory for the reverse SDE as $\mathcal{F}_h$ where $h>0$ is the step-size for numerical integration. Formally, we analyze the growth of $\Vert\mathbb{E}[\bar{\rvz}(t+h)] - \mathbb{E}[\bar{\rvz}_{t+h}]\Vert$ where $\bar{\rvz}_{t+h} = \rvz_{T-(t+h)} = \mathcal{G}_h(\bar{\rvz}_t)$ and $\bar{\rvz}(t+h) = \rvz(T-(t+h))= \mathcal{F}_h(\bar{\rvz}(t))$ are the approximated and ground-truth solutions at time $T-(t+h)$. This implies that we analyze convergence in a weak sense. However, this analysis is sufficient to present a justification for the proposed design choices for Reduced Splitting integrators. Furthermore, we have,
\begin{align}
    \bar{e}_{T-(t+h)} &= \Vert\mathbb{E}[\mathcal{F}_h(\bar{\rvz}(t))] - \mathbb{E}[\mathcal{G}_h(\bar{\rvz}_t)]\Vert \\
    &= \Vert\mathbb{E}[\mathcal{F}_h(\bar{\rvz}(t))] - \mathbb{E}[\mathcal{G}_h(\bar{\rvz}(t))] + \mathbb{E}[\mathcal{G}_h(\bar{\rvz}(t))] - \mathbb{E}[\mathcal{G}_h(\bar{\rvz}_t)]\Vert \\
    &\leq \Vert\mathbb{E}[\mathcal{F}_h(\bar{\rvz}(t))] - \mathbb{E}[\mathcal{G}_h(\bar{\rvz}(t))]\Vert + \Vert\mathbb{E}[\mathcal{G}_h(\bar{\rvz}(t))] - \mathbb{E}[\mathcal{G}_h(\bar{\rvz}_t)]\Vert
\end{align}
In a weak sense, the first term on the right-hand side of the above error bound is referred to as the \textit{local truncation error}. Intuitively, it gives an estimate of the numerical error introduced in the mean of the solution trajectory ($\mathbb{E}[\mathcal{F}_h(\bar{\rvz}(t))]$) due to our numerical scheme given the ground truth solution till the previous time step $t$. The second term in the error bound can be understood as the \textit{stability} of the numerical scheme. Intuitively, it gives an estimate of how much divergence is introduced by our numerical scheme given two nearby solution trajectories such that $\Vert\mathbb{E}[\rvz(t)] - \mathbb{E}[\rvz_t] \Vert < \delta$ for
some $\delta > 0$. Here, we only deal with the local truncation error in the position and the momentum space and leave stability analysis to future work. To this end, we first state the analytical form of the term $\mathbb{E}[\mathcal{F}_h(\rvz(t))]$.

\textbf{Computation of $\mathbb{E}[\mathcal{F}_h(\rvz(t))]$:} Using the Ito-Taylor expansion, it can be shown that the following results hold in the position and momentum space.
\begin{align}
    \mathbb{E}[\bar{\rvx}(t+h)] &= \mathbb{E}\Bigg[\bar{\rvx}(t) + \frac{h\beta}{2}\Bigg(\Gamma \bar{\rvx}(t) - M^{-1}\bar{\rvm}(t) + 2\Gamma\rvs_{\vtheta}^x\Bigg) + \frac{h^2\beta^2}{8}\Bigg[\Gamma\left(\Gamma \bar{\rvx}(t) - M^{-1}\bar{\rvm}(t) + 2\Gamma\rvs_{\vtheta}^x\right) \\
    & - M^{-1}\left(\bar{\rvx}(t) + \nu \bar{\rvm}(t) + 2M\nu \vs_{\vtheta}^m\right) + 2\Gamma\Big(\vs_{\vtheta}^{xx}(\Gamma \bar{\rvx}(t) - M^{-1}\bar{\rvm}(t) + 2\Gamma\rvs_{\vtheta}^x) + \\
    &\vs_{\vtheta}^{xm}(\bar{\rvx}(t) + \nu \bar{\rvm}(t) + 2M\nu \vs_{\vtheta}^m) + \frac{\partial \vs_{\vtheta}^x}{\partial t}\Big)\Bigg] + \mathcal{O}(h^3)\Bigg] \label{eqn:proof_0}
\end{align}

\begin{align}
     \mathbb{E}[\bar{\rvm}(t+h)] &=  \mathbb{E}\Bigg[\bar{\rvm}(t) + \frac{h\beta}{2}\Bigg(\bar{\rvx}(t) + \nu\bar{\rvm}(t) + 2M\nu\rvs_{\vtheta}^m\Bigg) + \frac{h^2\beta^2}{8}\Bigg[\left(\Gamma \bar{\rvx}(t) - M^{-1}\bar{\rvm}(t) + 2\Gamma\rvs_{\vtheta}^x\right) \\
    & + \nu\left(\bar{\rvx}(t) + \nu \bar{\rvm}(t) + 2M\nu \vs_{\vtheta}^m\right) + 2M\nu\Big(\vs_{\vtheta}^{mx}(\Gamma \bar{\rvx}(t) - M^{-1}\bar{\rvm}(t) + 2\Gamma\rvs_{\vtheta}^x) + \\
    &\vs_{\vtheta}^{mm}(\bar{\rvx}(t) + \nu \bar{\rvm}(t) + 2M\nu \vs_{\vtheta}^m) + \frac{\partial \vs_{\vtheta}^m}{\partial t}\Big)\Bigg] +  \mathcal{O}(h^3)\Bigg]
\end{align}
where $\vs_{\vtheta}^x$, $\vs_{\vtheta}^m$ denote the score components in the position and momentum space respectively. Moreover, $\vs_{\vtheta}^{xx}$, $\vs_{\vtheta}^{xm}$, $\vs_{\vtheta}^{mx}$ and $\vs_{\vtheta}^{mm}$ denote the partial derivatives $\frac{\partial \vs_{\vtheta}^x}{\partial x}$, $\frac{\partial \vs_{\vtheta}^x}{\partial m}$, $\frac{\partial \vs_{\vtheta}^m}{\partial x}$ and $\frac{\partial \vs_{\vtheta}^m}{\partial m}$, respectively.

We now illustrate the local truncation error (in a weak sense) for the NBAO sampler.

\subsection{Error Analysis: Naive BAO (NBAO)}
The NBAO sampler has the following update rules:
\begin{align}
    \bar{\rvm}_{t+h} &= \bar{\rvm}_t + \frac{h\beta}{2}\left[\bar{\rvx}_{t} +2\nu\bar{\rvm}_{t} + 2M\nu \vs_{\theta}^m(\bar{\rvx}_{t}, \bar{\rvm}_{t}, T - t)\right] \\
    \bar{\rvx}_{t+h} &= \bar{\rvx}_t + \frac{h\beta}{2}\left[2\Gamma \bar{\rvx}_t - M^{-1} \bar{\rvm}_{t+h} + 2\Gamma \vs_{\theta}^x(\bar{\rvx}_t, \bar{\rvm}_{t+h}, T - t)\right] \\
    \hat{\rvx}_{t+h} &= \exp{\left(\frac{-h\beta\Gamma}{2}\right)}\bar{\rvx}_{t+h} + \sqrt{1 - \exp{\left(-h\beta\Gamma\right)}}\bm{\epsilon}_x \label{eqn:proof_1}\\
    \hat{\rvm}_{t+h} &= \exp{\left(\frac{-h\beta\nu}{2}\right)}\bar{\rvm}_{t+h} + \sqrt{M}\sqrt{1 - \exp{\left(-h\beta\nu\right)}}\bm{\epsilon}_m \label{eqn:proof_2}
\end{align}

We compute the local truncation error for the NBAO sampler in both the position and the momentum space as follows.

\textbf{NBAO local truncation error in the position space:} From the update equations,
\begin{align}
    \bar{\rvx}(t+h) &= \bar{\rvx}(t) + \frac{h\beta}{2}\left[2\Gamma \bar{\rvx}(t) - M^{-1} \bar{\rvm}(t+h) + 2\Gamma \vs_{\theta}^x(\bar{\rvx}(t), \bar{\rvm}(t+h), T - t)\right] \\
    &= \bar{\rvx}(t) + \frac{h\beta}{2}\Bigg[2\Gamma \bar{\rvx}(t) - M^{-1} \Big(\bar{\rvm}(t) + \frac{h\beta}{2}\Big[\bar{\rvx}(t) + 2\nu\bar{\rvm}(t) + 2M\nu \vs_{\theta}^m(\bar{\rvx}(t), \bar{\rvm}(t), T - t)\Big]\Big)
\end{align}
\begin{align}
    &\qquad\qquad+ 2\Gamma \vs_{\theta}^x(\bar{\rvx}(t), \hat{\rvm}(t+h), T - t)\Bigg] \nonumber \\
    \bar{\rvx}(t+h) &= \bar{\rvx}(t) + \frac{h\beta}{2}\Bigg[2\Gamma \bar{\rvx}(t) - M^{-1}\bar{\rvm}(t) + 2\Gamma \vs_{\theta}^x(\bar{\rvx}(t), \hat{\rvm}(t+h), T - t)\Bigg] \\
    &\qquad\qquad-\frac{M^{-1}h^2\beta^2}{4}\Big[\bar{\rvx}(t) + 2\nu\bar{\rvm}(t) + 2M\nu \vs_{\theta}^m(\bar{\rvx}(t), \bar{\rvm}(t), T - t)\Big]\Big) \label{eqn:proof_3}
\end{align}
Lastly, we have the OU update in the position space, as follows:
\begin{align}
    \hat{\rvx}(t+h) &= \exp{\left(\frac{-h\beta\Gamma}{2}\right)}\bar{\rvx}(t+h) + \sqrt{1 - \exp{\left(-h\beta\Gamma\right)}}\bm{\epsilon}_x
\end{align}
We further approximate $\hat{\rvx}(t+h)$ from the update Eqn. \ref{eqn:proof_1} as follows,
\begin{align}
     \mathbb{E}[\hat{\rvx}(t+h)] &=  \mathbb{E}\Bigg[\Bigg[1 - \frac{\Gamma\beta h}{2} + \frac{\Gamma^2\beta^2 h^2}{8}\Bigg]\bar{\rvx}(t + h) + \mathcal{O}(h^3)\Bigg]
\end{align}
Substituting the expression in Eqn. \ref{eqn:proof_3} in the above approximation and ignoring higher order terms $\mathcal{O}(h^3)$, we have:
\begin{align}
    \mathbb{E}[\hat{\rvx}(t+h)] &= \mathbb{E}\Bigg[\bar{\rvx}(t + h) - \frac{\Gamma\beta h}{2}\bar{\rvx}(t + h) + \frac{\Gamma^2\beta^2 h^2}{8}\bar{\rvx}(t + h) + \mathcal{O}(h^3) \nonumber\\
    &=\bar{\rvx}(t) + \frac{h\beta}{2}\Bigg[2\Gamma\bar{\rvx}(t) - M^{-1}\bar{\rvm}(t) + 2\Gamma\vs_{\theta}^x(\bar{\rvx}(t), \bar{\rvm}(t+h), T - t)\Bigg] \nonumber\\
    &\quad- \frac{h^2\beta^2M^{-1}}{4}\Bigg[\bar{\rvx}(t) + 2\nu\bar{\rvm}(t) + 2M\nu \vs_{\theta}^m(\bar{\rvx}(t), \bar{\rvm}(t), T - t)\Bigg] - \frac{h\beta\Gamma}{2}\bar{\rvx}(t) \nonumber\\
    &\quad- \frac{h^2\beta^2\Gamma}{4}\Bigg[2\Gamma\bar{\rvx}(t) - M^{-1}\bar{\rvm}(t) + 2\Gamma\vs_{\theta}^x(\bar{\rvx}(t), \bar{\rvm}(t+h), T - t)\Bigg] + \frac{h^2\beta^2\Gamma^2}{8}\bar{\rvx}(t) + \mathcal{O}(h^3)\Bigg]\\
    \mathbb{E}[\hat{\rvx}(t+h)]&=\mathbb{E}\Bigg[\bar{\rvx}(t) + \frac{h\beta}{2}\Bigg[\Gamma\bar{\rvx}(t) - M^{-1}\bar{\rvm}(t) + 2\Gamma\vs_{\theta}^x(\bar{\rvx}(t), \bar{\rvm}(t+h), T - t)\Bigg] \nonumber\\
    &\quad- \frac{h^2\beta^2M^{-1}}{4}\Bigg[\bar{\rvx}(t) + 2\nu\bar{\rvm}(t) + 2M\nu \vs_{\theta}^m(\bar{\rvx}(t), \bar{\rvm}(t), T - t)\Bigg] \nonumber\\
    &\quad- \frac{h^2\beta^2\Gamma}{4}\Bigg[2\Gamma\bar{\rvx}(t) - M^{-1}\bar{\rvm}(t) + 2\Gamma\vs_{\theta}^x(\bar{\rvx}(t), \bar{\rvm}(t+h), T - t)\Bigg] + \frac{h^2\beta^2\Gamma^2}{8}\bar{\rvx}(t) + \mathcal{O}(h^3)\Bigg] \label{eqn:proof_4}
\end{align}

From Eqns. \ref{eqn:proof_0} and \ref{eqn:proof_4}, it follows that:
\begin{equation}
    \Bigg\Vert\mathbb{E}[\mathcal{F}_h(\bar{\rvz}(t))] - \mathbb{E}[\mathcal{G}_h(\bar{\rvz}(t))]\Bigg\Vert = \Bigg\Vert h\beta\Gamma\textcolor{blue}{\mathbb{E}\Big[\vs_{\theta}^x(\bar{\rvx}(t), \bar{\rvm}(t+h), T - t) - \vs_{\theta}^x(\bar{\rvx}(t), \bar
    {\rvm}(t), T - t)\Big]} + \mathcal{O}(h^2) \Bigg\Vert
\end{equation}

In the above equation, the error induced (in the position space) by the \textcolor{blue}{blue} term can be non-negligible, especially in the low step-size regimes (i.e. where the NFE is very small). Therefore, naive splitting integrators perform poorly in this case. A straightforward way to alleviate this problem is to re-use the score function evaluation $\vs_{\theta}(\bar{\rvx}_t, \bar{\rvm}_t, T-t)$ between consecutive updates for the position and the momentum space which leads to cancellation of this term, thereby preventing extra error terms from accumulating during sampling. This is precisely our first design choice in the formulation of reduced splitting integrators.

Similarly, our second choice of using timestep conditioning $T-(t+h)$ in the last step for the Reduced OBAB sampler is based on the error analysis in the momentum space for the NOBAB sampler. However, we omit the derivation here for simplicity.

Lastly, our choice of controlling the amount of noise injected in the position space is primarily inspired by EDM \citep{karraselucidating}. However, a theoretical analysis of this choice remains an interesting direction for future work.

\section{Extended Results}
\label{app:add_exps}
\subsection{Extended Results for Figure \ref{fig:fig_splitting}}
We present detailed results for the plots in Figure \ref{fig:fig_splitting} in Tables \ref{table:ext_1} and \ref{table:ext_2}.

\begin{table}[]
\footnotesize
\centering
\begin{tabular}{@{}ccccc@{}}
\toprule
Steps & EM SDE & NOBA & NBAO & NOBAB \\ \midrule
50    & 30.81  & 36.87     & 26.05     & 39.01      \\
70    & 15.63  & 24.23     & 16.98     & 17.2       \\
100   & 7.83   & 15.18     & 10.39     & 7.58       \\
150   & 4.26   & 9.68      & 6.44      & 3.71       \\
200   & 3.27   & 7.09      & 4.81      & 2.77       \\
250   & 2.75   & 5.56      & 4.15      & 2.64       \\
500   & 2.3    & 3.41      & 2.85      & 2.3        \\
1000  & 2.27   & 2.76      & 2.56      & 2.24       \\ \bottomrule
\end{tabular}
\caption{Extended Results for Figure \ref{fig:w1a}: Sample quality comparison between the Euler-Maruyama (EM) and Naive Splitting samplers on the CIFAR-10 dataset for different sampling budgets. FID (lower is better) reported on 50k samples.}
\label{table:ext_1}
\end{table}

\begin{table}[]
\footnotesize
\centering
\begin{tabular}{@{}ccccccc@{}}
\toprule
Steps & \multicolumn{2}{c}{OBA} & \multicolumn{2}{c}{BAO} & \multicolumn{2}{c}{OBAB} \\ \midrule
      & NOBA    & ROBA ($\lambda_s$)          & NBAO    & RBAO ($\lambda_s$)          & NOBAB    & ROBAB ($\lambda_s$)         \\ \midrule
50    & 36.87   & 2.76 (1.16)   & 26.05   & 3.33 (0.7)    & 39.01    & 6.86 (0.2)    \\
70    & 24.23   & 2.51 (0.66)   & 16.98   & 2.59 (0.44)   & 17.2     & 4.66 (0.16)   \\
100   & 15.18   & 2.36 (0.37)   & 10.39   & 2.65 (0.3)    & 7.58     & 3.54 (0.14)   \\
150   & 9.68    & 2.40 (0.2)    & 6.44    & 2.60 (0.18)   & 3.71     & 2.96 (0.12)   \\
200   & 7.09    & 2.38 (0.13)   & 4.81    & 2.43 (0.1)    & 2.77     & 2.67 (0.1)    \\ \bottomrule
\end{tabular}
\caption{Extended Results for Figures \ref{fig:w1b}-\ref{fig:w1d}: Sample quality comparison between naive and reduced splitting samplers on the CIFAR-10 dataset. Reduced versions outperform their naive counterparts for any given splitting scheme. The numbers in parentheses indicate the value of $\lambda_s$ used during sampling for the reduced integrators. FID (lower is better) reported on 50k samples.}
\label{table:ext_2}
\end{table}

\subsection{Additional Comparisons with SOTA SDE Solvers}
We present additional comparisons with state-of-the-art SDE solvers on CIFAR-10 in Table \ref{table:main}. Our proposed solver SPS (a.k.a Reduced OBA) outperforms competing baselines on similar compute budgets.
\begin{table}[]
\footnotesize
\centering
\begin{tabular}{@{}clcccc@{}}
\toprule
                                        &               &                                                             &           & \multicolumn{2}{c}{NFE (FID@50k $\downarrow$)} \\ \midrule
                                        & Method        & Description                                                 & Diffusion & 50                     & 100                   \\ \midrule
    &  (Ours) SPS & Splitting Integrator based PSLD Sampler    & PSLD      & \textbf{2.76}          & \textbf{2.36}  \\
                                        & SA-Solver \citep{xue2023sasolver}     & Stochastic Adams Solver applied to reverse SDEs             & VE        & 2.92                   & 2.63                  \\
                                        & SEEDS-2 \citep{gonzalez2023seeds}       & Exponential Integrators for SDEs (order=2)  & DDPM      & 11.10                  & 3.19                  \\
                                        & EDM \citep{karraselucidating}           & Custom stochastic sampler with churn                        & VP        & 3.19                   & 2.71                  \\
                                        & A-DDPM \citep{bao2022analyticdpm}        & Analytic variance estimation in reverse diffusion           & DDPM      & 5.50                   & 4.45                  \\
                                        & SSCS \citep{dockhornscore}          & Symmetric Splitting CLD Sampler                             & PSLD      & 18.83                  & 4.83                  \\
                                        & EM \citep{Kloeden1992}           & Euler Maruyama SDE sampler                                  & PSLD      & 30.81                  & 7.83                  \\ \bottomrule 
\end{tabular}
\caption{Our proposed stochastic samplers perform comparably or outperform prior methods for CIFAR-10. Diffusion: (VP,VE) \citep{songscore}, CLD \citep{dockhornscore}, DDPM \citep{ho2020denoising}, PSLD \citep{pandey2023generative}. Entries in \textbf{bold} indicate the best stochastic sampler for a given compute budget.}
\label{table:main}
\end{table}

\section{Implementation Details}
\label{app:exp_setup}
Here, we present complete implementation details for all the samplers presented in this work.

\subsection{Datasets and Preprocessing}
We use the CIFAR-10 \citep{krizhevsky2009learning} (50k images) and CelebA-64 (downsampled to 64 x 64 resolution, $\approx$ 200k images) \citep{liu2015faceattributes} datasets for both quantitative and qualitative analysis. During training, all datasets are preprocessed to a numerical range of [-1, 1]. Following prior work, we use random horizontal flips to train all new models across datasets as a data augmentation strategy. During inference, we re-scale all generated samples between the range [0, 1].

\subsection{Pre-trained Models}
For all ablation results in Section 3 in the main text, we use pre-trained PSLD \citep{pandey2023generative} models for CIFAR-10 with SDE hyperparameters $\Gamma=0.01$, $\nu=4.01$ and $\beta=8.0$. The resulting model consists of approximately 97M parameters. Similarly, we use a pre-trained PSLD model for CelebA-64 for state-of-the-art comparisons in Section 4. For more details on the score network architecture and training protocol, refer to \citet{pandey2023generative}.

\textbf{SDE Hyperparameters}: Similar to \citet{pandey2023generative}, we set $\beta=8.0$, $M^{-1}=4$ and $\gamma=0.04$ for all datasets. For CIFAR-10, we set $\Gamma=0.01$ and $\nu=4.01$, corresponding to the best settings in PSLD. Similarly, for CelebA-64, we set $\Gamma=0.005$ and $\nu=4.005$. Similar to \citet{pandey2023generative}, we add a stabilizing numerical epsilon value of $1e^{-9}$ in the diagonal entries of the Cholesky decomposition of $\mSigma_t$ when sampling from the perturbation kernel $p(\rvz_t|\rvx_0)$ during training.

\subsection{Evaluation} 
Unless specified otherwise, we report the FID \citep{heusel2017gans} score on 50k samples for assessing sample quality. Similarly, we use the network function evaluations (NFE) to assess sampling efficiency. In practice, we use the \texttt{torch-fidelity}\citep{obukhov2020torchfidelity} package for computing all FID reported in this work.

\textbf{Timestep Selection during Sampling}: We use quadratic striding for timestep discretization proposed in \citet{dockhornscore} during sampling, which ensures more number of score function evaluations in the lower timestep regime (i.e., $t$, which is close to the data). This kind of timestep selection is particularly useful when the NFE budget is limited. We also explored the timestep discretization proposed in \citet{karraselucidating} but noticed a degradation in sample quality.

\textbf{Last-Step Denoising}: It is common to add an Euler-based denoising step from a cutoff $\epsilon$ to zero to optimize for sample quality \citep{songscore, dockhornscore, jolicoeur-martineau2021adversarial} at the expense of another sampling step. For deterministic samplers presented in this work, we omit this heuristic due to observed degradation in sample quality. However, for stochastic samplers, we find that using last-step denoising leads to improvements in sample quality (especially when adjusting the amount of stochasticity as in Reduced Splitting Integrators). Formally, we perform the following update as a last denoising step for stochastic samplers:
\begin{align}
    \begin{pmatrix}
        \rvx_0 \\ \rvm_0 
    \end{pmatrix} = \begin{pmatrix}
        \rvx_{\eps} \\ \rvm_{\eps}
    \end{pmatrix} + \frac{\beta_t\eps}{2}\begin{pmatrix}
        \Gamma \rvx_{\eps} - M^{-1}\rvm_{\eps} + 2\Gamma \vs_{\theta}(\rvz_{\eps}, \eps)|_{0:d}\\
        \rvx_{\eps} + \nu \rvm_{\eps} + 2M\nu \vs_{\theta}(\rvz_{\eps}, \eps)|_{d:2d})
    \end{pmatrix}
\end{align}
Similar to PSLD, we set $\epsilon=1e-3$ during sampling for all experiments. Though recent works \citep{lu2022dpm, zhang2023fast} have found lower cutoffs to work better for a certain NFE budget, we leave this exploration in the context of PSLD to future work.

\end{document}